\PassOptionsToPackage{dvipsnames}{xcolor}
\documentclass[sigconf]{acmart}

\usepackage[most]{tcolorbox}

\usepackage{enumitem}

\usepackage{amssymb}
\usepackage{pifont}
\usepackage{tabularx}

\newcommand{\rqtag}[2]{%
\tikz[baseline=(rq.base)]\node[
    inner sep=2.0pt,
    rounded corners=2.5pt,
    draw=#1,
    line width=0.8pt,
    fill=#1!12,
    text=#1!85!black,
    font=\bfseries\footnotesize
](rq){RO};%
}

\definecolor{rq1}{HTML}{1F77B4} 

\newcommand{\cmark}{\textcolor{green!60!black}{\ding{51}}}
\newcommand{\xmark}{\textcolor{red!70!black}{\ding{55}}}

\newcommand*\circledblue[1]{%
\tikz[baseline=(char.base)]{
  \node[shape=circle, draw=NavyBlue!60, fill=NavyBlue!10, thick, inner sep=1pt] (char) {\scriptsize\textsf{#1}};
}}

\AtBeginDocument{%
  }

\copyrightyear{2026}
\acmYear{2026}
\setcopyright{cc}
\setcctype{by}
\acmConference[KDD '26]{Proceedings of the 32nd ACM SIGKDD Conference on Knowledge Discovery and Data Mining V.2}{August 09--13, 2026}{Jeju Island, Republic of Korea}
\acmBooktitle{Proceedings of the 32nd ACM SIGKDD Conference on Knowledge Discovery and Data Mining V.2 (KDD '26), August 09--13, 2026, Jeju Island, Republic of Korea}
\acmDOI{10.1145/3770855.3818647}
\acmISBN{979-8-4007-2259-2/2026/08}

\begin{document}

\title{Causal Methods for LLM Development and Evaluation}


\author{Dennis Frauen}
\author{Marie Brockschmidt}
\author{Konstantin Hess}
\author{Haorui Ma}
\author{Yuchen Ma}
\author{Abdurahman Maarouf}
\author{Maresa Schröder}
\author{Jonas Schweisthal}
\author{Yuxin Wang}
\affiliation{%
  \institution{MCML \& LMU Munich}
  \city{Munich}
  \country{Germany}
}


\author{Athiya Deviyani}
\affiliation{%
  \institution{CMU}
  \city{Pittsburgh}
  \country{USA}}
\email{adeviyan@cs.cmu.edu}

\author{Sonali Parbhoo}
\affiliation{%
  \institution{Imperial College London}
  \city{London}
  \country{UK}}
\email{s.parbhoo@imperial.ac.uk}

\author{Rahul G. Krishnan}
\affiliation{%
  \institution{University of Toronto}
  \city{Toronto}
  \country{Canada}}
\email{rahulgk@cs.toronto.edu}

\author{Stefan Feuerriegel}
\affiliation{%
  \institution{MCML \& LMU Munich}
  \city{Munich}
  \country{Germany}
}
\email{feuerriegel@lmu.de}

\renewcommand{\shortauthors}{Frauen et al.}

\makeatletter
\def\@mkauthors{%
  \global\setbox\mktitle@bx=\vbox{%
    \unvbox\mktitle@bx
    \par\medskip
    \begin{center}
      {\@authorfont
      \parbox{0.95\textwidth}{\centering
  \setlength{\baselineskip}{1.2\baselineskip}
        Dennis Frauen\textsuperscript{1},
        Marie Brockschmidt\textsuperscript{1},
        Konstantin Hess\textsuperscript{1},
        Haorui Ma\textsuperscript{1},
        Yuchen Ma\textsuperscript{1},
        Abdurahman Maarouf\textsuperscript{1},
        Maresa Schröder\textsuperscript{1},
        Jonas Schweisthal\textsuperscript{1},
        Yuxin Wang\textsuperscript{1},
        Athiya Deviyani\textsuperscript{2},
        Sonali Parbhoo\textsuperscript{3},
        Rahul G. Krishnan\textsuperscript{4},
        Stefan Feuerriegel\textsuperscript{1,*}\par}
      }

      \vspace{0.4em}

      {\@affiliationfont
        \textsuperscript{1}MCML \& LMU Munich, Munich, Germany, \textsuperscript{2}CMU, Pittsburgh, US, \textsuperscript{3}Imperial College London, London, UK\par 
        \textsuperscript{4}University of Toronto \& Vector Institute, Toronto, Canada, \textsuperscript{*}Correspondence: feuerriegel@lmu.de\par
      }
    \end{center}
    \par\bigskip
  }%
}
\makeatother

\begin{abstract}
Large language model (LLM) development is currently driven by large-scale empirical iteration over data mixtures, reward models, routing strategies, and evaluation pipelines. Here, we argue that many central questions in LLM development and evaluation are inherently causal: What is the effect of adding a data domain during pretraining? How do annotator preferences change when LLMs generate text in a different style? Should a prompt be routed to a larger or smaller model given inference cost constraints? In general, causal methods are well-suited to such settings where interventions change outcomes but, surprisingly, are underrepresented in LLM development. Our contribution is threefold: (1) We explain how causal methods can help develop modern LLM development and evaluation: LLM development relies heavily on logged data, which are often subject to confounding and distribution shifts; evaluation uses learned but potentially biased judges; and deployment environments are non-stationary. These conditions make purely predictive approaches fragile and create opportunities for principled identification and estimation methods from causal inference. (2) We further map opportunities for causal methods in the entire LLM development pipeline, including pretraining, alignment, routing, agentic workflows, and evaluation. (3) We discuss new research opportunities around leveraging causal methods for LLM development and evaluation. Overall, we argue that causal methods are potentially underutilized for the LLM development and evaluation pipeline, despite the fact that such methods can ensure a reliable and scientifically grounded design.
\end{abstract}




\maketitle

\section{Introduction}


The development of large language models (LLMs) involves many complex engineering decisions throughout the training and deployment pipeline \cite{grattafiori2024llama3herdmodels, guo2024large, touvron2023llama, zhao2023survey, huang2026probellm}. Engineers must decide, for instance, which data sources to include during pretraining, which model to route a query to, and how to evaluate competing systems under limited annotation budgets. Each such decision affects downstream performance, safety, and cost. Hence, developers must reason about counterfactual questions---that is, what was the effect of a given design decision, and what would have happened under an alternative choice---often without access to experimental evidence.


We argue that \emph{many decision problems in the LLM development pipeline can be framed through the lens of causal inference}.\footnote{For an overview of causal machine learning, see \citet{Kaddour.2022,Frauen2026MLCausal}.} In practice, developers frequently rely on historical logged data to guide design choices. However, such logged data are not generated under randomized experimentation, but originate from prior system configurations, human or AI feedback mechanisms, and deployment policies, and are therefore subject to selection bias, distribution shift, and noisy or biased labels \cite{Yao.2023}. LLMs further operate in evolving environments in which models, users, and tasks co-adapt over time, further widening the gap between past observations and future deployments \cite{packer2023memgpt, shinn2023reflexion}. As a result, it is often unclear whether observed performance differences are truly caused by a development decision or instead reflect confounding, evaluation artifacts, or spurious correlations. For example, if a routing system assigns harder queries to a larger model, a na{\"i}ve comparison of logged outcomes may incorrectly suggest that the larger model performs worse, when, in fact, it was simply evaluated on more difficult inputs \cite{Tsiourvas.2025}.

To this end, viewing decisions on LLM development and evaluation from a causal perspective has several \textbf{advantages}. \textbf{(1)}~Causal inference provides a principled framework for reasoning about the downstream effect of decisions in settings where controlled experimentation is costly, and where development instead relies on historical logged data (often called ``observational data'' in causal inference), yet which can be subject to selection bias, distribution shifts, and noisy or biased labels~\citep{shalit.2017estimating, Curth.2021nonparametric}. \textbf{(2)}~The use of causal techniques provides tools to address precisely such noise or biases introduced by feedback processes during human evaluation. In particular, causal methods can often help avoid spurious learning~\citep{Geirhos.2020} by isolating the effect of LLM development decisions rather than learning correlations that arise due to artifacts of the data collection or evaluation process (e.g., because human raters prefer longer responses). \textbf{(3)}~Causal methods can be used to increase data-efficiency, robustness against misspecification, and statistical validity (e.g., via valid confidence intervals)~\citep{Kennedy.2023, Kennedy.2023a, Chernozhukov.2018}.


In this paper, we propose a unifying perspective on leveraging causal methods in the LLM development and evaluation pipeline. \circledblue{1}~We explain \emph{why} and \emph{when} causal methods are well-suited to model problems along the LLM development and evaluation pipeline. In doing so, we identify common decisions in the pipeline as interventions whose effects on downstream utility, safety, and cost can be inferred from observational logs. \circledblue{2}~We map opportunities for causal methods across the LLM pipeline from pre-training, alignment, deployment, to evaluation, including LLM arenas, routing, LLM-as-a-judge, and agentic workflows. Building on this perspective, we propose practical principles for addressing the challenges that arise when causal inference meets large-scale LLM development and evaluation. \circledblue{3}~We outline impactful research directions for causal methods in LLM development, particularly for agentic systems and world models for which actions unfold through sequential decisions and interactions. In doing so, we draw on a growing research stream that applies causal methods to LLM development (e.g., \cite{Frauen.2026,Tsiourvas.2025,bou2026the,joselowitz2025insights,Guerdan.2026,Kobalczyk.2025}).

\begin{tcolorbox}[
  colback=ForestGreen!8,
  colframe=ForestGreen!75!black,
  boxrule=0.8pt,
  arc=1.5mm,
  left=1.2mm,
  right=1.2mm,
  top=1mm,
  bottom=1mm,
  title=\textbf{The scope of our Blue Sky vision},
  fonttitle=\bfseries,
  coltitle=white,
  enhanced,
  breakable
]
The intersection of causal reasoning and LLMs has received increasing attention, and, to clearly situate our vision, we distinguish how our scope is \textbf{different} from other research streams and thus novel: 
\medskip

\textbf{\xmark~LLMs to perform causal reasoning.}
This stream studies whether LLMs can answer causal questions and whether they possess causal knowledge about the world (e.g., \cite{jin2023can, jin2023cladder, kiciman2023causal, zevcevic2023causal}). Here, the idea is to equip LLMs with causal reasoning abilities. Put simply, causality is the \emph{output}.

\medskip
\textbf{\cmark~Causal methods for LLM development.}
Our focus is different in that causal methods are the \emph{input} for improving LLM development and evaluation. We ask under which settings and conditions decision tasks, typically made by developers rather than end users, can be framed as causal problems and addressed using causal methods.

\medskip
$\mathbf{\Rightarrow}$ In short, we do not ask whether LLMs can \emph{perform} causal reasoning; we ask how causal inference can help us \emph{build, evaluate, and optimize} LLM systems more reliably.
\end{tcolorbox}

\begin{tcolorbox}[
  colback=Red!5,
  colframe=Red!75!black,
  boxrule=0.8pt,
  arc=1.5mm,
  left=1mm,
  right=1mm,
  top=1mm,
  bottom=1mm,
  boxsep=1mm,
  title=\textbf{$\blacktriangle$ Disclaimer},
  fonttitle=\bfseries,
  enhanced,
  breakable
]
We caution against the expectation that simply incorporating ``causal methods'' into training objectives will automatically produce genuine causal reasoning abilities. Such hopes may be unwarranted unless the assumptions required for causal identification are explicitly satisfied \citep{Pearl.2009,Kern.2025}.
\end{tcolorbox}

\section{Why and when causal methods are helpful}

To explain why and when causal methods may be helpful, it is useful to start from the type of engineering questions that arise in the LLM development pipeline, where many decisions take the form: \emph{``If we change a system component, how will downstream behavior change?'' Causal inference provides a formal framework for answering precisely this type of question.} 

\textbf{Framing LLM development decisions as causal problems.}  Many development decisions in the LLM pipeline can be interpreted as causal interventions (e.g., modifying a data mixture, prompt template, router, retriever, or evaluation policy). The key quantity of interest is therefore unobserved and thus a \emph{counterfactual} quantity: what would have happened under a different intervention? At a high level, each system modification (such as changing a data mixture, adjusting alignment, selecting a routing rule, or altering retrieval depth) can be viewed as an intervention whose effect the engineer aims to estimate on downstream outcomes.

This decision task can be formalized as follows. Let $X$ denote the observed context (e.g., user prompt, pre-training corpus, retrieved documents), let $A$ represent a concrete system choice or treatment (e.g., prompt template, data mixture weights, retrieval policy, routing assignment), and let $Y$ denote an outcome of interest (e.g., task success, perplexity, preference score, safety metric, latency, or a combined utility). Using the potential outcomes framework \citep{Rubin.1974}, we write $Y(a)$ for the outcome that would be observed if the system were set to decision $A=a$. Often, the objective is to learn a causal quantity such as $\mathbb{E}[Y(a)]$ or differences $\mathbb{E}[Y(a)-Y(a')]$, quantifying the effect of alternative design choices. Importantly, this abstraction bridges causal terminology with standard NLP tasks: the same $(X,A,Y)$ structure applies to pretraining, alignment, deployment, and evaluation.

\textbf{Example.} \emph{Let us consider an LLM router that directs each user query to one of several candidate LLMs to balance response quality and inference cost \citep{Tsiourvas.2025}. For a given query with context $X$, the router selects a specific model $A$ (e.g., GPT-5.4 vs. Gemini), and the outcome $Y$ measures utility (e.g., a preference score minus latency cost). The developer has access to logged data with historical information about which model was assigned and the resulting outcome. However, queries may have been routed \emph{non-randomly}---for instance, harder queries may have been sent to larger models. A na{\"i}ve comparison of observed average outcomes across models thus does not identify the causal effect of assigning a query to a particular model. Instead, the quantity of interest is the counterfactual utility $\mathbb{E}[Y(a)]$, for example, with $a=\texttt{"gpt-5.4"}$, that would have been obtained if all queries had been routed to model $a$. In Section~\ref{sec:routing}, we show how we can recover this counterfactual utility from such confounded logs.}

\textbf{Principles of causal inference.} We now introduce two core principles of causal inference, namely, \circledblue{1}~\emph{identifiability}, which asks whether the causal effect of a decision can be recovered from observed data under explicit assumptions, and \circledblue{2}~\emph{estimability}, which concerns how such effects can be reliably and efficiently inferred from high-dimensional data. We then explain how these principles apply to LLM development and evaluation, and how they help clarify when decision effects can be learned from logged data.

\textbf{\circledblue{1} Identifiability.} Once development decisions are framed as causal interventions, a fundamental question arises: \emph{can the effect of such an intervention actually be recovered from the available data?} This is the problem of identifiability \citep{Pearl.2009}. Formally, identifiability asks whether the effect of interest is uniquely determined by the observed data under a set of explicit assumptions. In other words, do the logged data contain enough information, given clearly stated structural assumptions, to distinguish the true effect of a decision from alternative explanations? In the LLM pipeline, for instance, suppose evaluation relies solely on a learned LLM judge that is systematically biased toward long responses, and no unbiased human reference labels are available. In that case, the true effect of a design change on actual user utility is not identifiable from the observed judge scores alone. Increasing the number of judged examples only refines the biased estimate; it does not recover the true effect.

In the LLM pipeline, identifiability is challenging because the data is rarely generated under randomized experimentation: (i)~Logged data are typically \emph{policy-induced}, meaning that actions $A$ are chosen according to a historical policy rather than randomly assigned. (ii)~Outcomes may be \emph{selectively observed}, since only a subset of prompts or outputs receive labels. (iii)~Evaluation signals are often created by a judge instead of the end-user, and humans or learned judges can introduce measurement error or systematic bias. 

Causal modeling provides a formal language to reason about these issues \cite{Kern.2025}, and thus to clarify which \emph{assumptions} are required for identifiability: for example, whether conditioning on observed context $X$ suffices to remove confounding, whether missing labels can be treated as ignorable, or whether a judge score can be treated as a valid proxy for the desired outcome. It also highlights when such assumptions are implausible, and conclusions are therefore not supported by the data. Making these boundary conditions explicit is, as we argue later, a strength rather than a limitation: it allows practitioners to understand when an LLM decision problem can be reliably solved, and it provides methods to quantify potential biases, such as sensitivity analysis and partial identification \citep{Dorn.2024,Oprescu.2023,Frauen.2023c}. %

\textbf{\circledblue{2} Estimation}. Even when a causal effect is identifiable in principle, it must still be estimated reliably from finite and high-dimensional data. In LLM development, this challenge is particularly acute because available logged data are large but complex: the context $X$ is high-dimensional and may include raw text, conversation history, retrieved documents, user metadata, and model versions. At the same time, estimation typically requires learning additional auxiliary quantities---often called \emph{nuisance components}---such as the historical decision policy that generated the data, a reward model predicting human preferences, or a model predicting missing labels. These quantities are not the final object of interest, but they are necessary for adjusting bias and isolating the causal effect.

A key difficulty during inference is that errors in estimating these nuisance components can propagate into the final effect estimate, potentially rendering statistical inference invalid (e.g., invalid confidence intervals and suboptimal estimators). One remedy is based on \emph{semiparametric efficiency theory} \citep{Bickel.1998,vanderVaart.1998}, which motivates the construction of estimators via efficient influence functions that remove bias and yield valid statistical inference. The resulting methodology is commonly known as debiased/double machine learning (DML) \citep{Robins.1994,vanderLaan.2006,Chernozhukov.2018}. Intuitively, DML first predicts both the outcome and the treatment using any flexible ML model, and then estimates the causal effect on the residuals, thereby removing confounding.

In the remainder of this paper, we apply the causal lens centered on identifiability and estimation to the LLM development pipeline. For each stage, we examine how identifiability determines what can be learned from logged data and how causal methods can be used for efficient estimation.

\section{Causal methods for LLM development}\label{sec:pipeline}

\begin{figure}[h] 
    \centering
    \includegraphics[width=1\linewidth]{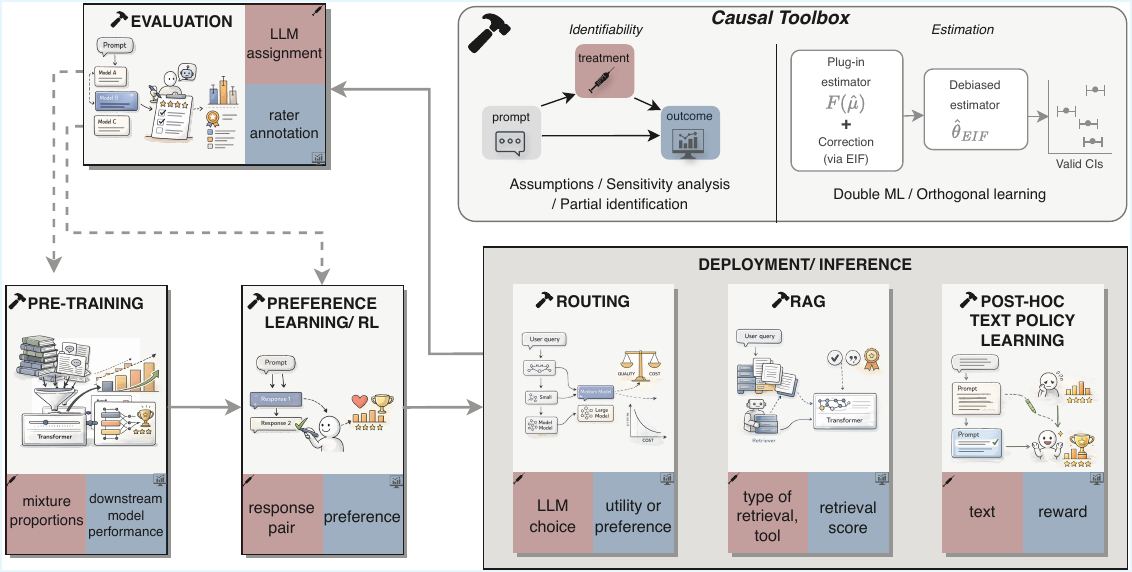}
    \vspace{-0.7cm}
    \caption{Overview of using causal methods for LLM development and evaluation.}
    \Description{Overview of using causal methods for LLM development and evaluation.}
    \label{fig:framework}
    \vspace{-0.4cm}
\end{figure}

We now walk through the LLM development pipeline (see Fig.~\ref{fig:framework}), map problems that can be modeled via causal inference, and outline research opportunities (marked as \emph{\rqtag{rq1}{1}}).

\vspace{-0.2cm}
\subsection{LLM pre-training} 

Pre-training datasets are mixtures of sources (e.g., books, code, forums) \cite{Gao.2020-ThePile}. The domain weights, i.e., the proportion of sampling of each domain, strongly affect the training efficiency and downstream model performance \cite{Du.2022}. However, these choices can introduce \emph{selection bias}: toxicity filtering may disproportionately remove dialectal text \cite{Sap.2019, Dixon.2018}, English-centric filtering may shift topic distributions \cite{Dodge.2021, Bender.2021}, and deduplication may alter the frequency of facts and the balance between memorization and generalization \cite{Lee.2022, Carlini.2021, Kandpal.2022}. From a causal perspective, these curation choices can be viewed as interventions on the training distribution, with mixture proportions as treatments $A$, the data sources and training context as covariates $X$, and downstream model performance as the outcome $Y$. In practice, fully retraining models for every candidate mixture is often infeasible. As a result, developers rely on heuristics or small-scale experiments to guide dataset curation, often subject to selection bias.  \emph{\textbf{$\Rightarrow$ \rqtag{rq1}{1}:} This suggests learning a model that predicts downstream performance as a function of corpus weights, enabling developers to optimize pre-training composition without fully retraining on every candidate mixture. Methodologically, this connects to continuous treatment effect estimation (e.g., \cite{Kennedy.2017}). Similarly, orthogonalization-based techniques can be used for variance reduction in hyperparameter optimization \citep{Schroeder.2026}.}

\vspace{-0.2cm}
\subsection{LLM alignment and preference learning}

The goal of LLM alignment is to better align a pretrained model with human preferences \citep{Ouyang.2022}. In practice, alignment often includes at least two stages \citep{Ouyang.2022}. First, the model is instruction fine-tuned on carefully curated prompt-response pairs. Second, the model is further adapted using preference learning, where the objective is to match human preferences over alternative outputs. 

Causal methods are particularly useful for the preference learning step. In a standard setup for preference learning \citep{Kallus.2025}, an LLM generates two candidate responses $A = (A_1, A_2)$ conditional on a prompt $X$. An annotator indicates a preference $Y$ over the responses. From a causal perspective, the prompt can be viewed as the context or confounder $X$, the presented response pair as the treatment $A$, and the observed preference as the outcome $Y$. This framing clarifies how prompts, candidate responses, and annotator feedback jointly determine the observed preference data.

A key challenge is that preference data only provide relative comparisons, while latent utility scores or rewards remain unobserved. The dominant approach is the Bradley--Terry (BT) model \citep{Bradley.1952}, which underlies both RLHF \citep{Christiano.2017} (where a reward model is learned from pairwise comparisons and used for policy optimization) and DPO \citep{Rafailov.2023} (which directly learns an optimal policy from BT).

However, the BT model can be misspecified in several ways. For example, pairwise preferences may exhibit cyclic patterns that cannot be represented by a single latent score function \citep{Xu.2025}. More broadly, insights from social choice theory suggest that preference data generally need not admit a globally optimal score representation \citep{Zhang.2024}. As a result, methods that rely too strongly on the BT model may lead to biased or unstable conclusions when the assumed structure is violated. This is a natural point of entry for causal and double machine learning (DML) methodology. Recent work has proposed doubly robust alignment procedures based on DML for win-rate style targets \citep{Xu.2025b}, as well as extensions incorporating additional signals such as response time \citep{Sawarni.2025}. At the same time, several semiparametric extensions of the BT model have been developed to improve robustness to misspecification \citep{Frauen.2026, Kallus.2025, Li.2025, Spokoiny.2025}. Overall, preference learning can benefit from the same principles that have proven successful in causal inference, such as orthogonalization, nuisance-robust estimation, and semiparametric efficiency.

A causal viewpoint is also valuable for understanding how preference data are collected and where biases may arise. In practice, the data-generating process for alignment is often selective and may be misspecified if important assumptions are ignored, making identifiability challenging \citep{Kobalczyk.2025}. For example, some responses are more likely to be shown to annotators than others. Unobserved confounding can be especially problematic when responses or annotations are missing not at random. For instance, \citet{Xia.2024} propose an instrumental-variable framework to remove confounding biases during alignment. \emph{\textbf{$\Rightarrow$ \rqtag{rq1}{1}:} More generally, this raises a broader research agenda: tools from causal inference such as sensitivity analysis \citep{Frauen.2023c}, partial identification, and explicit missing-data modeling may provide principled ways to reason about preference data when standard assumptions are violated.}

\vspace{-0.5cm}
\subsection{LLM deployment}

\subsubsection{LLM routing}
\label{sec:routing}

LLM routing refers to the problem of assigning each user query $X$ (here the confounder) to a model $A$ (treatment) from a set of candidate models to optimize some utility of interest $Y$~\citep{Ding.2024, Hu.2024} (outcome). LLM routers are trained on data generated under an existing routing or evaluation policy: each query is assigned to a model, and the outcomes under alternative model assignments remain unobserved~\cite{Tsiourvas.2025}. Alternatively, in preference datasets, only a sparse, policy-dependent subset of model pairs is ever compared~\citep{Chiang.2024, Frick.2025}. As a result, observed outcomes are confounded by the historical assignment mechanism. Causal methods are essential in clarifying when identification from observational data is possible, and in developing robust adjustment and off-policy learning methods. Recent work on counterfactual estimation and regret minimization~\citep{Tsiourvas.2025} shows that accounting for treatment bias can substantially improve routing performance. \emph{\textbf{$\Rightarrow$ \rqtag{rq1}{1}:} The question of whether the prompt and logged context suffice for backdoor adjustment is open. Similar to the preference learning setting, other (partial) identification approaches may be explored as alternatives.}

\vspace{-0.1cm}
\subsubsection{Multi-agent workflows and RAG}
Modern LLM deployments are agentic pipelines composed of multiple interacting components \citep{zhou2024webarena, park2023generative, wang2024voyager}. Examples include multi-model routing \citep{Zhang.2025, Hu.2024, Frick.2025} and tool-using agents \citep{schick2023toolformer, Yao.2023}. From a causal perspective, these systems can be viewed as sequential decision processes: given a user query $Q$, the system takes actions $A_1,\ldots,A_T$ (e.g., retrieval, routing, or tool invocation) that produce intermediate outputs $Z_1,\ldots,Z_T$ and determine a downstream outcome $Y$. Many system design questions therefore correspond to causal estimands, such as the effect of increasing retrieval depth, invoking additional tools, or routing a query to a larger model. Retrieval augmented generation (RAG) \citep{Lewis.2020, Gao.2024, Karpukhin.2020, Izacard.2021} provides a concrete instance of such pipelines: the system retrieves documents conditioned on the query and then generates an answer from them. Retrieved evidence thus acts as an intermediate variable mediating the final outcome. This creates a credit assignment problem across system components, since errors may arise from the retriever, the generator, or their interaction \cite{Leung.2026}. From a causal perspective, this mediation structure complicates analysis because retrieved evidence affects both generation quality and computational cost, and alternative retrieval policies may produce document sets not observed in logged data.

Agentic systems are typically trained using observational logs collected under existing policies, which introduces confounding due to factors such as query difficulty, user group, or routing strategies \cite{Tsiourvas.2025}. Unlike single-shot prediction settings, agents actively interact with their environment: their actions influence subsequent states, observations, and ultimately long-term rewards. This interaction induces rich cause-effect structures, where early decisions shape downstream utilities and constraints. Such sequential decision-making problems are complicated in terms of identifiability and estimation because earlier system choices influence later observations and final outcomes, leading naturally to off-policy evaluation and policy learning problems \citep{Wu.2026, Dudik.2011, Kausik.2024}. \emph{\textbf{$\Rightarrow$ \rqtag{rq1}{1}:} Identifying and evaluating strategic behavior of agents (i.e., sequential decision policies with long-term rewards) remains an open challenge. Estimating the causal effect of actions requires off-policy evaluation methods that handle structured mediators (e.g., retrieved documents or tool outputs), account for dynamic confounding across time, and remain reliable under limited support and costly end-to-end feedback.}

\vspace{-0.1cm}
\subsubsection{Offline policy learning with text/prompt optimization.}

LLMs are often used to choose or generate text itself, such as prompts, rewrites, recommendations, explanations, or safety instructions. This yields a deployment-time problem that is distinct from RLHF or fine-tuning: the model can be fixed, while the system optimizes a textual action from logged interactions. This imposes a typical causal inference setup, where the confounders $X$ are observed context, such as conversation history, user features, or retrieved documents; the treatment $A$ is the chosen text artifact; and the outcome $Y$ is a downstream outcome, such as user satisfaction, task success, or click-rate. The goal is then to learn a policy $\pi(X)$ that selects $A$ to maximize $\mathbb{E}[Y(\pi)]$ \cite{feder2022causal, Veitch.2020}.

In practice, historical texts are generated under an existing policy and depend on confounders such as user difficulty, prior interaction state, or safety heuristics. As a result, na{\"i}ve prediction of observed reward does not identify the counterfactual utility of alternative texts. This makes the problem a natural instance of off-policy evaluation and policy learning with structured, high-dimensional actions \cite{saito2022off, saito2023off, saitopotec2025}. Prompt optimization fits the same formulation, with the prompt as treatment and the final outcome as reward \citep{kiyohara2025off}. Such methods from logged bandits, doubly robust or orthogonal policy learning \cite{athey2021policy}, and treatment-effect estimation with structured actions are therefore directly relevant \cite{kramer2026causal, kaddour2021causal}. \emph{\textbf{$\Rightarrow$ \rqtag{rq1}{1}:} A key open question is identification under weak overlap or unobserved confounding, since many candidate texts or prompts are observed only for a narrow subset of contexts and observed rewards may be influenced by hidden confounders.}

\vspace{-0.4cm}
\subsection{LLM evaluation}

The goal of LLM evaluation is to assess and rank models on target tasks such as coding, writing, or reasoning. Framing this task in causal terms, the prompt is the confounder $X$, the LLM assignment is the treatment $A$, and the annotation is the outcome $Y$. 

\textbf{Preference ranking.} Estimating ranking scores is closely related to treatment effect estimation: the central question is how outcomes differ across model assignments after accounting for prompt heterogeneity. Moreover, automated metrics themselves can vary across contexts, and relying on them in lieu of human annotations can lead to different model rankings \citep{Deviyani.2025}. \emph{\textbf{$\Rightarrow$ \rqtag{rq1}{1}:} DML provides a principled way to adjust for these contextual shifts, enabling valid estimation and confidence intervals even when nuisance components are learned with flexible black-box models \citep{Frauen.2026}. Furthermore, it can address BT model misspecification analogously to the alignment setting and has also been applied to evaluation under distribution shift \citep{Ulichney.2025}. Rigorous evaluation further requires uncertainty quantification to determine whether observed performance differences are statistically meaningful, and annotations may come from crowdsourced annotators, as in LMArena \citep{Chiang.2024, Chi.2025, Miroyan.2025}, or domain experts for specialized tasks \citep{Bedi.2025}.}

\textbf{Bias assessment.} Biases arise when prompts are not assigned uniformly. For example, an LLM router sending harder prompts to larger models makes prompt difficulty a confounder, systematically underestimating the larger model's true performance. Similarly, selective missingness in annotations can distort rankings. This can come from instances where annotators are less likely to rate especially difficult or lengthy examples. Causal inference provides a toolkit for quantifying these biases explicitly. \emph{\textbf{$\Rightarrow$ \rqtag{rq1}{1}:} At the same time, it remains largely open how to adapt causal methods such as sensitivity analysis to the specific structure of LLM evaluation.}

\textbf{LLM-as-a-judge.} LLM judges offer scale and low cost but introduce systematic biases (position effects, verbosity preferences, self-enhancement) \cite{koo:ar-bias, park2024offsetbias, wang-etal-2024-large-language-models-fair}. Methods inspired by DML and prediction-powered inference (PPI) \citep{Angelopoulos.2023, van2026calibeating} address this by combining large-scale LLM annotations with a small gold-standard human sample to obtain debiased estimates with formal guarantees \citep{Chatzi.2024, Fisch.2024, Frauen.2026, Guerdan.2026}. \emph{\textbf{$\Rightarrow$ \rqtag{rq1}{1}:} Key open questions are how large these gains are in practice and how to select the best judge model for a given task and population.}

\textbf{Active data collection.} Causal methodology also informs data collection design when human feedback is costly: DML-based design methods provide variance-minimizing annotation allocation rules \citep{Cook.2024,Zrnic.2024}, with extensions to cost-optimal active evaluation \citep{Angelopoulos.2025}, informative pairwise comparison selection \citep{Jamieson.2011}, and optimal design under BT models \citep{Guo.2018, Frauen.2026}. \emph{\textbf{$\Rightarrow$ \rqtag{rq1}{1}:} An especially promising direction for future work is the design of optimal evaluation schemes for complex sequential systems such as LLM agents, where obtaining labels may require evaluating the full costly system pipeline.}

\textbf{Agent evaluation.} For agentic systems, final outcome metrics such as pass/fail scores are often insufficient for evaluation, because they can obscure benchmark artifacts, intermediate steps such as routing decisions and tool calls, failure modes, and safety-relevant behavior during execution. \emph{\textbf{$\Rightarrow$ \rqtag{rq1}{1}:} More research is needed on moving from final-outcome evaluation to causal evaluation of agent behavior, including how interventions on intermediate components such as tools, retrieval policies, or action-selection rules affect downstream utility and safety. One example is log analysis, i.e., the systematic tracking of inputs, execution traces, tool use, and outputs \citep{Kirgis.2026}), which can help diagnose issues and has many synergies with causal modeling, as it exposes intermediate variables needed to reason about mediation, dynamic confounding, and failure attribution in sequential LLM systems.}

\vspace{-0.1cm}
\subsection{Safety audits}

When a model produces toxic, misaligned, or otherwise harmful outputs, the relevant question is \emph{why}, i.e., which aspect of the training or alignment pipeline \emph{caused} the failure. For example, harmful behavior may arise from biases in the pretraining data \citep{Feng.2023}, misspecified reward models during alignment \citep{Mire.2025}, or other unintended optimization incentives. Diagnosing and correcting safety failures therefore requires external tools that reason about the latent reward functions driving behavior (e.g., learned via RLHF).

Formally, the challenge is whether the underlying reward or objective function is identified, that is, whether it can be uniquely inferred from the model's observable behavior \citep{joselowitz2025insights}. In general, this is difficult, because multiple reward functions can produce the same distribution of outputs. For example, two reward functions may produce identical outputs during training while encouraging very different behaviors under a distribution shift. As a result, behavioral observations alone may be insufficient to identify the true objective that the model is optimizing. One approach is to treat reward discovery as a (Bayesian) partial identification problem to explicitly yield a distribution over plausible reward functions \citep{bou2026the}. \emph{\textbf{$\Rightarrow$ \rqtag{rq1}{1}:} A key open question is when latent reward functions are identifiable from behavioral observations and how identifiability degrades under distribution shift or adversarial inputs. Adapting sensitivity analysis and partial identification tools to characterize the set of reward functions consistent with observed behavior, rather than inferring a single point estimate, would allow developers to bound the probability of harmful actions under unseen deployment conditions and formally trace alignment failures back to their source in the pipeline.}

\vspace{-0.1cm}
\section{Conclusion}

We argue that many central decisions in modern LLM development can be framed as causal tasks. Looking forward, bridging causal inference and LLM development will likely require new methodological advances. We thus view causal machine learning as a promising foundation for reliable, data-efficient, and deployment-aware LLM systems, and we are confident this perspective helps motivate a broader LLM research agenda toward reliable large-scale AI systems.

\clearpage
\begin{acks}
This work has been supported by the German Federal Ministry of Education and Research (Grant: 01IS24082). This paper is supported by the DAAD program ``Konrad Zuse Schools of Excellence in Artificial Intelligence'', sponsored by the Federal Ministry of Education and Research.
\end{acks}

\bibliographystyle{ACM-Reference-Format}
\bibliography{literature}


\begin{thebibliography}{105}


\ifx \showCODEN    \undefined \def \showCODEN     #1{\unskip}     \fi
\ifx \showISBNx    \undefined \def \showISBNx     #1{\unskip}     \fi
\ifx \showISBNxiii \undefined \def \showISBNxiii  #1{\unskip}     \fi
\ifx \showISSN     \undefined \def \showISSN      #1{\unskip}     \fi
\ifx \showLCCN     \undefined \def \showLCCN      #1{\unskip}     \fi
\ifx \shownote     \undefined \def \shownote      #1{#1}          \fi
\ifx \showarticletitle \undefined \def \showarticletitle #1{#1}   \fi
\ifx \showURL      \undefined \def \showURL       {\relax}        \fi
\providecommand\bibfield[2]{#2}
\providecommand\bibinfo[2]{#2}
\providecommand\natexlab[1]{#1}
\providecommand\showeprint[2][]{arXiv:#2}

\bibitem[Angelopoulos et~al\mbox{.}(2023)]%
        {Angelopoulos.2023}
\bibfield{author}{\bibinfo{person}{Anastasios~N. Angelopoulos}, \bibinfo{person}{Stephen Bates}, \bibinfo{person}{Clara Fannjiang}, \bibinfo{person}{Michael~I. Jordan}, {and} \bibinfo{person}{Tijana Zrnic}.} \bibinfo{year}{2023}\natexlab{}.
\newblock \showarticletitle{Prediction-powered inference}.
\newblock \bibinfo{journal}{\emph{Science}} \bibinfo{volume}{382}, \bibinfo{number}{6671} (\bibinfo{year}{2023}), \bibinfo{pages}{669--674}.
\newblock


\bibitem[Angelopoulos et~al\mbox{.}(2025)]%
        {Angelopoulos.2025}
\bibfield{author}{\bibinfo{person}{Anastasios~N. Angelopoulos}, \bibinfo{person}{Jacob Eisenstein}, \bibinfo{person}{Jonathan Berant}, \bibinfo{person}{Alekh Agarwal}, {and} \bibinfo{person}{Adam Fisch}.} \bibinfo{year}{2025}\natexlab{}.
\newblock \showarticletitle{Cost-optimal active {{AI}} model evaluation}.
\newblock   \bibinfo{volume}{arXiv:2506.07949} (\bibinfo{year}{2025}).
\newblock


\bibitem[Athey and Wager(2021)]%
        {athey2021policy}
\bibfield{author}{\bibinfo{person}{Susan Athey} {and} \bibinfo{person}{Stefan Wager}.} \bibinfo{year}{2021}\natexlab{}.
\newblock \showarticletitle{Policy learning with observational data}.
\newblock \bibinfo{journal}{\emph{Econometrica}} \bibinfo{volume}{89}, \bibinfo{number}{1} (\bibinfo{year}{2021}), \bibinfo{pages}{133--161}.
\newblock


\bibitem[Bedi et~al\mbox{.}(2026)]%
        {Bedi.2025}
\bibfield{author}{\bibinfo{person}{Suhana Bedi}, \bibinfo{person}{Hejie Cui}, \bibinfo{person}{Miguel Fuentes}, \bibinfo{person}{Alyssa Unell}, \bibinfo{person}{Michael Wornow}, \bibinfo{person}{Juan~M Banda}, \bibinfo{person}{Nikesh Kotecha}, \bibinfo{person}{Timothy Keyes}, \bibinfo{person}{Yifan Mai}, \bibinfo{person}{Mert Oez}, {et~al\mbox{.}}} \bibinfo{year}{2026}\natexlab{}.
\newblock \showarticletitle{Holistic evaluation of large language models for medical tasks with MedHELM}.
\newblock \bibinfo{journal}{\emph{Nature Medicine}}  \bibinfo{volume}{32} (\bibinfo{year}{2026}), \bibinfo{pages}{943--951}.
\newblock


\bibitem[Bender et~al\mbox{.}(2021)]%
        {Bender.2021}
\bibfield{author}{\bibinfo{person}{Emily~M. Bender}, \bibinfo{person}{Timnit Gebru}, \bibinfo{person}{Angelina McMillan-Major}, {and} \bibinfo{person}{Shmargaret Shmitchell}.} \bibinfo{year}{2021}\natexlab{}.
\newblock \showarticletitle{On the dangers of stochastic parrots: Can language models be too big?}. In \bibinfo{booktitle}{\emph{FAccT}}.
\newblock


\bibitem[Bickel et~al\mbox{.}(1998)]%
        {Bickel.1998}
\bibfield{author}{\bibinfo{person}{Peter~J. Bickel}, \bibinfo{person}{Chris A.~J. Klaassen}, \bibinfo{person}{Ya'acov Ritov}, {and} \bibinfo{person}{Jon~A. Wellner}.} \bibinfo{year}{1998}\natexlab{}.
\newblock \bibinfo{booktitle}{\emph{Efficient and adaptive estimation for semiparametric models}}.
\newblock \bibinfo{publisher}{Springer, New York}.
\newblock
\showISBNx{978-0-387-98473-5}


\bibitem[Bou et~al\mbox{.}(2026)]%
        {bou2026the}
\bibfield{author}{\bibinfo{person}{Matthieu Bou}, \bibinfo{person}{Nyal Patel}, \bibinfo{person}{Arjun Jagota}, \bibinfo{person}{Satyapriya Krishna}, {and} \bibinfo{person}{Sonali Parbhoo}.} \bibinfo{year}{2026}\natexlab{}.
\newblock \showarticletitle{The alignment auditor: A Bayesian framework for verifying and refining {LLM} objectives}. In \bibinfo{booktitle}{\emph{ICLR}}.
\newblock


\bibitem[Bradley and Terry(1952)]%
        {Bradley.1952}
\bibfield{author}{\bibinfo{person}{Ralph~Allan Bradley} {and} \bibinfo{person}{Milton~E. Terry}.} \bibinfo{year}{1952}\natexlab{}.
\newblock \showarticletitle{Rank analysis of incomplete block designs: {{I}}. {{The}} method of paired comparisons}.
\newblock \bibinfo{journal}{\emph{Biometrika}} \bibinfo{volume}{39}, \bibinfo{number}{3/4} (\bibinfo{year}{1952}), \bibinfo{pages}{324--345}.
\newblock


\bibitem[Carlini et~al\mbox{.}(2021)]%
        {Carlini.2021}
\bibfield{author}{\bibinfo{person}{Nicholas Carlini}, \bibinfo{person}{Florian Tramer}, \bibinfo{person}{Eric Wallace}, \bibinfo{person}{Matthew Jagielski}, \bibinfo{person}{Ariel Herbert-Voss}, \bibinfo{person}{Katherine Lee}, \bibinfo{person}{Adam Roberts}, \bibinfo{person}{Tom Brown}, \bibinfo{person}{Dawn Song}, \bibinfo{person}{Ulfar Erlingsson}, {et~al\mbox{.}}} \bibinfo{year}{2021}\natexlab{}.
\newblock \showarticletitle{Extracting training data from large language models}. In \bibinfo{booktitle}{\emph{USENIX Security Symposium}}.
\newblock


\bibitem[Chatzi et~al\mbox{.}(2024)]%
        {Chatzi.2024}
\bibfield{author}{\bibinfo{person}{Ivi Chatzi}, \bibinfo{person}{Eleni Straitouri}, \bibinfo{person}{Suhas Thejaswi}, {and} \bibinfo{person}{Manuel~Gomez Rodriguez}.} \bibinfo{year}{2024}\natexlab{}.
\newblock \showarticletitle{Prediction-powered ranking of large language models}. In \bibinfo{booktitle}{\emph{{{NeurIPS}}}}.
\newblock


\bibitem[Chernozhukov et~al\mbox{.}(2018)]%
        {Chernozhukov.2018}
\bibfield{author}{\bibinfo{person}{Victor Chernozhukov}, \bibinfo{person}{Denis Chetverikov}, \bibinfo{person}{Mert Demirer}, \bibinfo{person}{Esther Duflo}, \bibinfo{person}{Christian Hansen}, \bibinfo{person}{Whitney Newey}, {and} \bibinfo{person}{James~M. Robins}.} \bibinfo{year}{2018}\natexlab{}.
\newblock \showarticletitle{Double/debiased machine learning for treatment and structural parameters}.
\newblock \bibinfo{journal}{\emph{The Econometrics Journal}} \bibinfo{volume}{21}, \bibinfo{number}{1} (\bibinfo{year}{2018}), \bibinfo{pages}{1--68}.
\newblock


\bibitem[Chi et~al\mbox{.}(2025)]%
        {Chi.2025}
\bibfield{author}{\bibinfo{person}{Wayne Chi}, \bibinfo{person}{Valerie Chen}, \bibinfo{person}{Anastasios~Nikolas Angelopoulos}, \bibinfo{person}{Wei-Lin Chiang}, \bibinfo{person}{Aditya Mittal}, \bibinfo{person}{Naman Jain}, \bibinfo{person}{Tianjun Zhang}, \bibinfo{person}{Ion Stoica}, \bibinfo{person}{Chris Donahue}, {and} \bibinfo{person}{Ameet Talwalkar}.} \bibinfo{year}{2025}\natexlab{}.
\newblock \showarticletitle{{Copilot Arena}: A platform for code {LLM} evaluation in the wild}. In \bibinfo{booktitle}{\emph{ICML}}.
\newblock


\bibitem[Chiang et~al\mbox{.}(2024)]%
        {Chiang.2024}
\bibfield{author}{\bibinfo{person}{Wei-Lin Chiang}, \bibinfo{person}{Lianmin Zheng}, \bibinfo{person}{Ying Sheng}, \bibinfo{person}{Anastasios~N Angelopoulos}, \bibinfo{person}{Tianle Li}, \bibinfo{person}{Dacheng Li}, \bibinfo{person}{Banghua Zhu}, \bibinfo{person}{Hao Zhang}, \bibinfo{person}{Michael~I Jordan}, \bibinfo{person}{Joseph~E Gonzalez}, {and} \bibinfo{person}{Ion Stoica}.} \bibinfo{year}{2024}\natexlab{}.
\newblock \showarticletitle{Chatbot {{Arena}}: {{An}} open platform for evaluating {{LLMs}} by human preference}. In \bibinfo{booktitle}{\emph{{{ICML}}}}.
\newblock


\bibitem[Christiano et~al\mbox{.}(2017)]%
        {Christiano.2017}
\bibfield{author}{\bibinfo{person}{Paul~F Christiano}, \bibinfo{person}{Jan Leike}, \bibinfo{person}{Tom Brown}, \bibinfo{person}{Miljan Martic}, \bibinfo{person}{Shane Legg}, {and} \bibinfo{person}{Dario Amodei}.} \bibinfo{year}{2017}\natexlab{}.
\newblock \showarticletitle{Deep reinforcement learning from human preferences}. In \bibinfo{booktitle}{\emph{{{NeurIPS}}}}.
\newblock


\bibitem[Cook et~al\mbox{.}(2024)]%
        {Cook.2024}
\bibfield{author}{\bibinfo{person}{Thomas Cook}, \bibinfo{person}{Alan Mishler}, {and} \bibinfo{person}{Aaditya Ramdas}.} \bibinfo{year}{2024}\natexlab{}.
\newblock \showarticletitle{Semiparametric efficient inference in adaptive experiments}. In \bibinfo{booktitle}{\emph{{{CLeaR}}}}.
\newblock


\bibitem[Curth and Van~der Schaar(2021)]%
        {Curth.2021nonparametric}
\bibfield{author}{\bibinfo{person}{Alicia Curth} {and} \bibinfo{person}{Mihaela Van~der Schaar}.} \bibinfo{year}{2021}\natexlab{}.
\newblock \showarticletitle{Nonparametric estimation of heterogeneous treatment effects: From theory to learning algorithms}. In \bibinfo{booktitle}{\emph{AISTATS}}.
\newblock


\bibitem[Deviyani and Diaz(2025)]%
        {Deviyani.2025}
\bibfield{author}{\bibinfo{person}{Athiya Deviyani} {and} \bibinfo{person}{Fernando Diaz}.} \bibinfo{year}{2025}\natexlab{}.
\newblock \showarticletitle{Contextual metric meta-evaluation by measuring local metric accuracy}. In \bibinfo{booktitle}{\emph{Findings of the ACL: NAACL 2025}}.
\newblock


\bibitem[Ding et~al\mbox{.}(2024)]%
        {Ding.2024}
\bibfield{author}{\bibinfo{person}{Dujian Ding}, \bibinfo{person}{Ankur Mallick}, \bibinfo{person}{Chi Wang}, \bibinfo{person}{Robert Sim}, \bibinfo{person}{Subhabrata Mukherjee}, \bibinfo{person}{Victor Ruhle}, \bibinfo{person}{Laks V.~S. Lakshmanan}, {and} \bibinfo{person}{Ahmed~Hassan Awadallah}.} \bibinfo{year}{2024}\natexlab{}.
\newblock \showarticletitle{Hybrid LLM: Cost-efficient and quality-aware query routing}. In \bibinfo{booktitle}{\emph{ICLR}}.
\newblock


\bibitem[Dixon et~al\mbox{.}(2018)]%
        {Dixon.2018}
\bibfield{author}{\bibinfo{person}{Lucas Dixon}, \bibinfo{person}{John Li}, \bibinfo{person}{Jeffrey Sorensen}, \bibinfo{person}{Nithum Thain}, {and} \bibinfo{person}{Lucy Vasserman}.} \bibinfo{year}{2018}\natexlab{}.
\newblock \showarticletitle{Measuring and mitigating unintended bias in text classification}. In \bibinfo{booktitle}{\emph{AIES}}.
\newblock


\bibitem[Dodge et~al\mbox{.}(2021)]%
        {Dodge.2021}
\bibfield{author}{\bibinfo{person}{Jesse Dodge}, \bibinfo{person}{Maarten Sap}, \bibinfo{person}{Ana Marasovi{\'c}}, \bibinfo{person}{William Agnew}, \bibinfo{person}{Gabriel Ilharco}, \bibinfo{person}{Dirk Groeneveld}, \bibinfo{person}{Margaret Mitchell}, {and} \bibinfo{person}{Matt Gardner}.} \bibinfo{year}{2021}\natexlab{}.
\newblock \showarticletitle{Documenting large webtext corpora: A case study on the colossal clean crawled corpus}. In \bibinfo{booktitle}{\emph{EMNLP}}.
\newblock


\bibitem[Dorn et~al\mbox{.}(2025)]%
        {Dorn.2024}
\bibfield{author}{\bibinfo{person}{Jacob Dorn}, \bibinfo{person}{Kevin Guo}, {and} \bibinfo{person}{Nathan Kallus}.} \bibinfo{year}{2025}\natexlab{}.
\newblock \showarticletitle{Doubly-valid/doubly-sharp sensitivity analysis for causal inference with unmeasured confounding}.
\newblock \bibinfo{journal}{\emph{Journal of the American Statistical Association}} \bibinfo{volume}{120}, \bibinfo{number}{549} (\bibinfo{year}{2025}), \bibinfo{pages}{331--342}.
\newblock


\bibitem[Du et~al\mbox{.}(2022)]%
        {Du.2022}
\bibfield{author}{\bibinfo{person}{Nan Du}, \bibinfo{person}{Yanping Huang}, \bibinfo{person}{Andrew~M Dai}, \bibinfo{person}{Simon Tong}, \bibinfo{person}{Dmitry Lepikhin}, \bibinfo{person}{Yuanzhong Xu}, \bibinfo{person}{Maxim Krikun}, \bibinfo{person}{Yanqi Zhou}, \bibinfo{person}{Adams~Wei Yu}, \bibinfo{person}{Orhan Firat}, \bibinfo{person}{Barret Zoph}, \bibinfo{person}{Liam Fedus}, \bibinfo{person}{Maarten~P Bosma}, \bibinfo{person}{Zongwei Zhou}, \bibinfo{person}{Tao Wang}, \bibinfo{person}{Emma Wang}, \bibinfo{person}{Kellie Webster}, \bibinfo{person}{Marie Pellat}, \bibinfo{person}{Kevin Robinson}, \bibinfo{person}{Kathleen Meier-Hellstern}, \bibinfo{person}{Toju Duke}, \bibinfo{person}{Lucas Dixon}, \bibinfo{person}{Kun Zhang}, \bibinfo{person}{Quoc Le}, \bibinfo{person}{Yonghui Wu}, \bibinfo{person}{Zhifeng Chen}, {and} \bibinfo{person}{Claire Cui}.} \bibinfo{year}{2022}\natexlab{}.
\newblock \showarticletitle{{GL}a{M}: Efficient scaling of language models with mixture-of-experts}. In \bibinfo{booktitle}{\emph{ICML}}.
\newblock


\bibitem[Dud\'{\i}k et~al\mbox{.}(2011)]%
        {Dudik.2011}
\bibfield{author}{\bibinfo{person}{Miroslav Dud\'{\i}k}, \bibinfo{person}{John Langford}, {and} \bibinfo{person}{Lihong Li}.} \bibinfo{year}{2011}\natexlab{}.
\newblock \showarticletitle{Doubly robust policy evaluation and learning}. In \bibinfo{booktitle}{\emph{ICML}}.
\newblock


\bibitem[Feder et~al\mbox{.}(2022)]%
        {feder2022causal}
\bibfield{author}{\bibinfo{person}{Amir Feder}, \bibinfo{person}{Katherine~A Keith}, \bibinfo{person}{Emaad Manzoor}, \bibinfo{person}{Reid Pryzant}, \bibinfo{person}{Dhanya Sridhar}, \bibinfo{person}{Zach Wood-Doughty}, \bibinfo{person}{Jacob Eisenstein}, \bibinfo{person}{Justin Grimmer}, \bibinfo{person}{Roi Reichart}, \bibinfo{person}{Margaret~E Roberts}, {et~al\mbox{.}}} \bibinfo{year}{2022}\natexlab{}.
\newblock \showarticletitle{Causal inference in natural language processing: Estimation, prediction, interpretation and beyond}. In \bibinfo{booktitle}{\emph{TACL}}.
\newblock


\bibitem[Feng et~al\mbox{.}(2023)]%
        {Feng.2023}
\bibfield{author}{\bibinfo{person}{Shangbin Feng}, \bibinfo{person}{Chan~Young Park}, \bibinfo{person}{Yuhan Liu}, {and} \bibinfo{person}{Yulia Tsvetkov}.} \bibinfo{year}{2023}\natexlab{}.
\newblock \showarticletitle{From pretraining data to language models to downstream tasks: Tracking the trails of political biases leading to unfair {NLP} models}. In \bibinfo{booktitle}{\emph{ACL}}.
\newblock


\bibitem[Fisch et~al\mbox{.}(2024)]%
        {Fisch.2024}
\bibfield{author}{\bibinfo{person}{Adam Fisch}, \bibinfo{person}{Joshua Maynez}, \bibinfo{person}{R~Alex Hofer}, \bibinfo{person}{Bhuwan Dhingra}, \bibinfo{person}{Amir Globerson}, {and} \bibinfo{person}{William~W Cohen}.} \bibinfo{year}{2024}\natexlab{}.
\newblock \showarticletitle{Stratified prediction-powered inference for hybrid language model evaluation}. In \bibinfo{booktitle}{\emph{{{NeurIPS}}}}.
\newblock


\bibitem[Frauen et~al\mbox{.}(2026a)]%
        {Frauen.2026}
\bibfield{author}{\bibinfo{person}{Dennis Frauen}, \bibinfo{person}{Athiya Deviyani}, \bibinfo{person}{Mihaela van~der Schaar}, {and} \bibinfo{person}{Stefan Feuerriegel}.} \bibinfo{year}{2026}\natexlab{a}.
\newblock \showarticletitle{Nonparametric {{LLM}} evaluation from preference data}. In \bibinfo{booktitle}{\emph{ICML}}.
\newblock


\bibitem[Frauen et~al\mbox{.}(2023)]%
        {Frauen.2023c}
\bibfield{author}{\bibinfo{person}{Dennis Frauen}, \bibinfo{person}{Valentyn Melnychuk}, {and} \bibinfo{person}{Stefan Feuerriegel}.} \bibinfo{year}{2023}\natexlab{}.
\newblock \showarticletitle{Sharp bounds for generalized causal sensitivity analysis}. In \bibinfo{booktitle}{\emph{{{NeurIPS}}}}.
\newblock


\bibitem[Frauen et~al\mbox{.}(2026b)]%
        {Frauen2026MLCausal}
\bibfield{author}{\bibinfo{person}{Dennis Frauen}, \bibinfo{person}{Valentyn Melnychuk}, \bibinfo{person}{Lars van~der Laan}, {and} \bibinfo{person}{Stefan Feuerriegel}.} \bibinfo{year}{2026}\natexlab{b}.
\newblock \showarticletitle{Machine Learning for Causal Inference}. In \bibinfo{booktitle}{\emph{Wiley StatsRef: Statistics Reference Online}}. \bibinfo{publisher}{John Wiley \& Sons}.
\newblock


\bibitem[Frick et~al\mbox{.}(2025)]%
        {Frick.2025}
\bibfield{author}{\bibinfo{person}{Evan Frick}, \bibinfo{person}{Connor Chen}, \bibinfo{person}{Joseph Tennyson}, \bibinfo{person}{Tianle Li}, \bibinfo{person}{Wei-Lin Chiang}, \bibinfo{person}{Anastasios~N. Angelopoulos}, {and} \bibinfo{person}{Ion Stoica}.} \bibinfo{year}{2025}\natexlab{}.
\newblock \showarticletitle{Prompt-to-leaderboard: {{Prompt-adaptive LLM}} evaluations}. In \bibinfo{booktitle}{\emph{{{ICML}}}}.
\newblock


\bibitem[Gao et~al\mbox{.}(2020)]%
        {Gao.2020-ThePile}
\bibfield{author}{\bibinfo{person}{Leo Gao}, \bibinfo{person}{Stella Biderman}, \bibinfo{person}{Sid Black}, \bibinfo{person}{Laurence Golding}, \bibinfo{person}{Travis Hoppe}, \bibinfo{person}{Charles Foster}, \bibinfo{person}{Jason Phang}, \bibinfo{person}{Horace He}, \bibinfo{person}{Anish Thite}, \bibinfo{person}{Noa Nabeshima}, \bibinfo{person}{Shawn Presser}, {and} \bibinfo{person}{Connor Leahy}.} \bibinfo{year}{2020}\natexlab{}.
\newblock \bibinfo{title}{The pile: An 800GB dataset of diverse text for language modeling}.
\newblock
\showeprint[arxiv]{2101.00027}


\bibitem[Gao et~al\mbox{.}(2024)]%
        {Gao.2024}
\bibfield{author}{\bibinfo{person}{Yunfan Gao}, \bibinfo{person}{Yun Xiong}, \bibinfo{person}{Xinyu Gao}, \bibinfo{person}{Kangxiang Jia}, \bibinfo{person}{Jinliu Pan}, \bibinfo{person}{Yuxi Bi}, \bibinfo{person}{Yi Dai}, \bibinfo{person}{Jiawei Sun}, \bibinfo{person}{Meng Wang}, {and} \bibinfo{person}{Haofen Wang}.} \bibinfo{year}{2024}\natexlab{}.
\newblock \bibinfo{title}{Retrieval-augmented generation for large language models: A survey}.
\newblock
\showeprint[arxiv]{2312.10997}


\bibitem[Geirhos et~al\mbox{.}(2020)]%
        {Geirhos.2020}
\bibfield{author}{\bibinfo{person}{Robert Geirhos}, \bibinfo{person}{J{\"o}rn-Henrik Jacobsen}, \bibinfo{person}{Claudio Michaelis}, \bibinfo{person}{Richard Zemel}, \bibinfo{person}{Wieland Brendel}, \bibinfo{person}{Matthias Bethge}, {and} \bibinfo{person}{Felix~A Wichmann}.} \bibinfo{year}{2020}\natexlab{}.
\newblock \showarticletitle{Shortcut learning in deep neural networks}.
\newblock \bibinfo{journal}{\emph{Nature Machine Intelligence}} \bibinfo{volume}{2}, \bibinfo{number}{11} (\bibinfo{year}{2020}), \bibinfo{pages}{665--673}.
\newblock


\bibitem[Grattafiori et~al\mbox{.}(2024)]%
        {grattafiori2024llama3herdmodels}
\bibfield{author}{\bibinfo{person}{Aaron Grattafiori}, \bibinfo{person}{Abhimanyu Dubey}, \bibinfo{person}{Abhinav Jauhri}, {et~al\mbox{.}}} \bibinfo{year}{2024}\natexlab{}.
\newblock \bibinfo{title}{The {Llama 3} herd of models}.
\newblock
\showeprint[arxiv]{2407.21783}


\bibitem[Guerdan et~al\mbox{.}(2026)]%
        {Guerdan.2026}
\bibfield{author}{\bibinfo{person}{Luke Guerdan}, \bibinfo{person}{Justin Whitehouse}, \bibinfo{person}{Kimberly Truong}, \bibinfo{person}{Ken Holstein}, {and} \bibinfo{person}{Steven Wu}.} \bibinfo{year}{2026}\natexlab{}.
\newblock \showarticletitle{Doubly-robust {LLM}-as-a-judge: Externally valid estimation with imperfect personas}. In \bibinfo{booktitle}{\emph{ICLR}}.
\newblock


\bibitem[Guo et~al\mbox{.}(2024)]%
        {guo2024large}
\bibfield{author}{\bibinfo{person}{Taicheng Guo}, \bibinfo{person}{Xiuying Chen}, \bibinfo{person}{Yaqi Wang}, \bibinfo{person}{Ruidi Chang}, \bibinfo{person}{Shichao Pei}, \bibinfo{person}{Nitesh~V Chawla}, \bibinfo{person}{Olaf Wiest}, {and} \bibinfo{person}{Xiangliang Zhang}.} \bibinfo{year}{2024}\natexlab{}.
\newblock \showarticletitle{Large language model based multi-agents: A survey of progress and challenges}. In \bibinfo{booktitle}{\emph{IJCAI}}.
\newblock


\bibitem[Guo et~al\mbox{.}(2018)]%
        {Guo.2018}
\bibfield{author}{\bibinfo{person}{Yuan Guo}, \bibinfo{person}{Peng Tian}, \bibinfo{person}{Jayashree {Kalpathy-Cramer}}, \bibinfo{person}{Susan Ostmo}, \bibinfo{person}{J.Peter Campbell}, \bibinfo{person}{Michael F.Chiang}, \bibinfo{person}{Deniz Erdogmus}, \bibinfo{person}{Jennifer Dy}, {and} \bibinfo{person}{Stratis Ioannidis}.} \bibinfo{year}{2018}\natexlab{}.
\newblock \showarticletitle{Experimental design under the {Bradley-Terry} model}. In \bibinfo{booktitle}{\emph{{{IJCAI}}}}.
\newblock


\bibitem[Hu et~al\mbox{.}(2024)]%
        {Hu.2024}
\bibfield{author}{\bibinfo{person}{Qitian~Jason Hu}, \bibinfo{person}{Jacob Bieker}, \bibinfo{person}{Xiuyu Li}, \bibinfo{person}{Nan Jiang}, \bibinfo{person}{Benjamin Keigwin}, \bibinfo{person}{Gaurav Ranganath}, \bibinfo{person}{Kurt Keutzer}, {and} \bibinfo{person}{Shriyash~Kaustubh Upadhyay}.} \bibinfo{year}{2024}\natexlab{}.
\newblock \showarticletitle{{RouterBench}: A benchmark for multi-{LLM} routing system}. In \bibinfo{booktitle}{\emph{Agentic Markets Workshop at ICML 2024}}.
\newblock


\bibitem[Huang et~al\mbox{.}(2026)]%
        {huang2026probellm}
\bibfield{author}{\bibinfo{person}{Yue Huang}, \bibinfo{person}{Zhengzhe Jiang}, \bibinfo{person}{Yuchen Ma}, \bibinfo{person}{Yu Jiang}, \bibinfo{person}{Xiangqi Wang}, \bibinfo{person}{Yujun Zhou}, \bibinfo{person}{Yuexing Hao}, \bibinfo{person}{Kehan Guo}, \bibinfo{person}{Pin-Yu Chen}, \bibinfo{person}{Stefan Feuerriegel}, {et~al\mbox{.}}} \bibinfo{year}{2026}\natexlab{}.
\newblock \showarticletitle{Probellm: Automating Principled Diagnosis of LLM Failures}. In \bibinfo{booktitle}{\emph{ICML}}.
\newblock


\bibitem[Izacard and Grave(2021)]%
        {Izacard.2021}
\bibfield{author}{\bibinfo{person}{Gautier Izacard} {and} \bibinfo{person}{Edouard Grave}.} \bibinfo{year}{2021}\natexlab{}.
\newblock \showarticletitle{Leveraging passage retrieval with generative models for open domain question answering}. In \bibinfo{booktitle}{\emph{EACL}}.
\newblock


\bibitem[Jamieson and Nowak(2011)]%
        {Jamieson.2011}
\bibfield{author}{\bibinfo{person}{Kevin~G Jamieson} {and} \bibinfo{person}{Robert Nowak}.} \bibinfo{year}{2011}\natexlab{}.
\newblock \showarticletitle{Active ranking using pairwise comparisons}. In \bibinfo{booktitle}{\emph{NeurIPS}}.
\newblock


\bibitem[Jin et~al\mbox{.}(2023)]%
        {jin2023cladder}
\bibfield{author}{\bibinfo{person}{Zhijing Jin}, \bibinfo{person}{Yuen Chen}, \bibinfo{person}{Felix Leeb}, \bibinfo{person}{Luigi Gresele}, \bibinfo{person}{Ojasv Kamal}, \bibinfo{person}{LYU Zhiheng}, \bibinfo{person}{Kevin Blin}, \bibinfo{person}{Fernando~Gonzalez Adauto}, \bibinfo{person}{Max Kleiman-Weiner}, \bibinfo{person}{Mrinmaya Sachan}, {et~al\mbox{.}}} \bibinfo{year}{2023}\natexlab{}.
\newblock \showarticletitle{{CLadder:} Assessing causal reasoning in language models}. In \bibinfo{booktitle}{\emph{NeurIPS}}.
\newblock


\bibitem[Jin et~al\mbox{.}(2024)]%
        {jin2023can}
\bibfield{author}{\bibinfo{person}{Zhijing Jin}, \bibinfo{person}{Jiarui Liu}, \bibinfo{person}{Zhiheng Lyu}, \bibinfo{person}{Spencer Poff}, \bibinfo{person}{Mrinmaya Sachan}, \bibinfo{person}{Rada Mihalcea}, \bibinfo{person}{Mona Diab}, {and} \bibinfo{person}{Bernhard Sch{\"o}lkopf}.} \bibinfo{year}{2024}\natexlab{}.
\newblock \showarticletitle{Can large language models infer causation from correlation?}. In \bibinfo{booktitle}{\emph{ICLR}}.
\newblock


\bibitem[Joselowitz et~al\mbox{.}(2025)]%
        {joselowitz2025insights}
\bibfield{author}{\bibinfo{person}{Jared Joselowitz}, \bibinfo{person}{Ritam Majumdar}, \bibinfo{person}{Arjun Jagota}, \bibinfo{person}{Matthieu Bou}, \bibinfo{person}{Nyal Patel}, \bibinfo{person}{Satyapriya Krishna}, {and} \bibinfo{person}{Sonali Parbhoo}.} \bibinfo{year}{2025}\natexlab{}.
\newblock \showarticletitle{Insights from the inverse: reconstructing {LLM} training goals through inverse reinforcement learning}. In \bibinfo{booktitle}{\emph{Second Conference on Language Modeling}}.
\newblock


\bibitem[Kaddour et~al\mbox{.}(2025)]%
        {Kaddour.2022}
\bibfield{author}{\bibinfo{person}{Jean Kaddour}, \bibinfo{person}{Aengus Lynch}, \bibinfo{person}{Qi Liu}, \bibinfo{person}{Matt~J. Kusner}, {and} \bibinfo{person}{Ricardo Silva}.} \bibinfo{year}{2025}\natexlab{}.
\newblock \showarticletitle{Causal Machine learning: A survey and open problems}.
\newblock \bibinfo{journal}{\emph{Foundations and Trends in Optimization}}  \bibinfo{volume}{9} (\bibinfo{year}{2025}), \bibinfo{pages}{1--247}.
\newblock
Issue 1-2.


\bibitem[Kaddour et~al\mbox{.}(2021)]%
        {kaddour2021causal}
\bibfield{author}{\bibinfo{person}{Jean Kaddour}, \bibinfo{person}{Yuchen Zhu}, \bibinfo{person}{Qi Liu}, \bibinfo{person}{Matt~J Kusner}, {and} \bibinfo{person}{Ricardo Silva}.} \bibinfo{year}{2021}\natexlab{}.
\newblock \showarticletitle{Causal effect inference for structured treatments}. In \bibinfo{booktitle}{\emph{NeurIPS}}.
\newblock


\bibitem[Kallus(2025)]%
        {Kallus.2025}
\bibfield{author}{\bibinfo{person}{Nathan Kallus}.} \bibinfo{year}{2025}\natexlab{}.
\newblock \showarticletitle{Semiparametric preference optimization: {{Your}} language model is secretly a single-index model}.
\newblock \bibinfo{journal}{\emph{arXiv preprint}}  \bibinfo{volume}{arXiv:2512.21917} (\bibinfo{year}{2025}).
\newblock


\bibitem[Kandpal et~al\mbox{.}(2022)]%
        {Kandpal.2022}
\bibfield{author}{\bibinfo{person}{Nikhil Kandpal}, \bibinfo{person}{Eric Wallace}, {and} \bibinfo{person}{Colin Raffel}.} \bibinfo{year}{2022}\natexlab{}.
\newblock \showarticletitle{Deduplicating training data mitigates privacy risks in language models}. In \bibinfo{booktitle}{\emph{ICML}}.
\newblock


\bibitem[Karpukhin et~al\mbox{.}(2020)]%
        {Karpukhin.2020}
\bibfield{author}{\bibinfo{person}{Vladimir Karpukhin}, \bibinfo{person}{Barlas Oguz}, \bibinfo{person}{Sewon Min}, \bibinfo{person}{Patrick Lewis}, \bibinfo{person}{Ledell Wu}, \bibinfo{person}{Sergey Edunov}, \bibinfo{person}{Danqi Chen}, {and} \bibinfo{person}{Wen-tau Yih}.} \bibinfo{year}{2020}\natexlab{}.
\newblock \showarticletitle{Dense passage retrieval for open-domain question answering}. In \bibinfo{booktitle}{\emph{EMNLP}}.
\newblock


\bibitem[Kausik et~al\mbox{.}(2024)]%
        {Kausik.2024}
\bibfield{author}{\bibinfo{person}{Chinmaya Kausik}, \bibinfo{person}{Yangyi Lu}, \bibinfo{person}{Kevin Tan}, \bibinfo{person}{Maggie Makar}, \bibinfo{person}{Yixin Wang}, {and} \bibinfo{person}{Ambuj Tewari}.} \bibinfo{year}{2024}\natexlab{}.
\newblock \showarticletitle{Offline policy evaluation and optimization under confounding}. In \bibinfo{booktitle}{\emph{AISTATS}}.
\newblock


\bibitem[Kennedy(2023a)]%
        {Kennedy.2023a}
\bibfield{author}{\bibinfo{person}{Edward~H. Kennedy}.} \bibinfo{year}{2023}\natexlab{a}.
\newblock \showarticletitle{Semiparametric doubly robust targeted double machine learning: a review}.
\newblock \bibinfo{journal}{\emph{2203.06469}} (\bibinfo{year}{2023}).
\newblock


\bibitem[Kennedy(2023b)]%
        {Kennedy.2023}
\bibfield{author}{\bibinfo{person}{Edward~H. Kennedy}.} \bibinfo{year}{2023}\natexlab{b}.
\newblock \showarticletitle{Towards optimal doubly robust estimation of heterogeneous causal effects}.
\newblock \bibinfo{journal}{\emph{Electronic Journal of Statistics}} \bibinfo{volume}{17}, \bibinfo{number}{2} (\bibinfo{year}{2023}), \bibinfo{pages}{3008--3049}.
\newblock


\bibitem[Kennedy et~al\mbox{.}(2017)]%
        {Kennedy.2017}
\bibfield{author}{\bibinfo{person}{Edward~H Kennedy}, \bibinfo{person}{Zongming Ma}, \bibinfo{person}{Matthew~D McHugh}, {and} \bibinfo{person}{Dylan~S Small}.} \bibinfo{year}{2017}\natexlab{}.
\newblock \showarticletitle{Non-parametric methods for doubly robust estimation of continuous treatment effects}.
\newblock \bibinfo{journal}{\emph{Journal of the Royal Statistical Society Series {B}: Statistical Methodology}} \bibinfo{volume}{79}, \bibinfo{number}{4} (\bibinfo{year}{2017}), \bibinfo{pages}{1229--1245}.
\newblock


\bibitem[Kern et~al\mbox{.}(2025)]%
        {Kern.2025}
\bibfield{author}{\bibinfo{person}{Christoph Kern}, \bibinfo{person}{Unai Fischer-Abaigar}, \bibinfo{person}{Jonas Schweisthal}, \bibinfo{person}{Dennis Frauen}, \bibinfo{person}{Rayid Ghani}, \bibinfo{person}{Stefan Feuerriegel}, \bibinfo{person}{Mihaela van~der Schaar}, {and} \bibinfo{person}{Frauke Kreuter}.} \bibinfo{year}{2025}\natexlab{}.
\newblock \showarticletitle{Algorithms for reliable decision-making need causal reasoning}.
\newblock \bibinfo{journal}{\emph{Nature Computational Science}} \bibinfo{volume}{5}, \bibinfo{number}{5} (\bibinfo{year}{2025}), \bibinfo{pages}{356--360}.
\newblock


\bibitem[K{\i}c{\i}man et~al\mbox{.}(2023)]%
        {kiciman2023causal}
\bibfield{author}{\bibinfo{person}{Emre K{\i}c{\i}man}, \bibinfo{person}{Robert Ness}, \bibinfo{person}{Amit Sharma}, {and} \bibinfo{person}{Chenhao Tan}.} \bibinfo{year}{2023}\natexlab{}.
\newblock \showarticletitle{Causal reasoning and large language models: Opening a new frontier for causality}.
\newblock \bibinfo{journal}{\emph{TMLR}} (\bibinfo{year}{2023}).
\newblock


\bibitem[Kirgis et~al\mbox{.}(2026)]%
        {Kirgis.2026}
\bibfield{author}{\bibinfo{person}{Peter Kirgis}, \bibinfo{person}{Sayash Kapoor}, \bibinfo{person}{Stephan Rabanser}, \bibinfo{person}{Nitya Nadgir}, \bibinfo{person}{Cozmin Ududec}, \bibinfo{person}{Magda Dubois}, \bibinfo{person}{JJ Allaire}, \bibinfo{person}{Conrad Stosz}, \bibinfo{person}{Marius Hobbhahn}, \bibinfo{person}{Jacob Steinhardt}, {et~al\mbox{.}}} \bibinfo{year}{2026}\natexlab{}.
\newblock \showarticletitle{Log analysis is necessary for credible evaluation of {AI} agents}.
\newblock \bibinfo{journal}{\emph{arXiv:2605.08545}} (\bibinfo{year}{2026}).
\newblock


\bibitem[Kiyohara et~al\mbox{.}(2025)]%
        {kiyohara2025off}
\bibfield{author}{\bibinfo{person}{Haruka Kiyohara}, \bibinfo{person}{Daniel~Yiming Cao}, \bibinfo{person}{Yuta Saito}, {and} \bibinfo{person}{Thorsten Joachims}.} \bibinfo{year}{2025}\natexlab{}.
\newblock \showarticletitle{An Off-Policy Learning Approach for Steering Sentence Generation towards Personalization}. In \bibinfo{booktitle}{\emph{RecSys}}.
\newblock


\bibitem[Kobalczyk and van~der Schaar(2025)]%
        {Kobalczyk.2025}
\bibfield{author}{\bibinfo{person}{Kasia Kobalczyk} {and} \bibinfo{person}{Mihaela van~der Schaar}.} \bibinfo{year}{2025}\natexlab{}.
\newblock \showarticletitle{Preference learning for {AI} alignment: A causal perspective}. In \bibinfo{booktitle}{\emph{ICML}}.
\newblock


\bibitem[Koo et~al\mbox{.}(2024)]%
        {koo:ar-bias}
\bibfield{author}{\bibinfo{person}{Ryan Koo}, \bibinfo{person}{Minhwa Lee}, \bibinfo{person}{Vipul Raheja}, \bibinfo{person}{Jong~Inn Park}, \bibinfo{person}{Zae~Myung Kim}, {and} \bibinfo{person}{Dongyeop Kang}.} \bibinfo{year}{2024}\natexlab{}.
\newblock \showarticletitle{Benchmarking cognitive biases in large language models as evaluators}. In \bibinfo{booktitle}{\emph{Findings of the ACL: ACL 2024}}.
\newblock


\bibitem[Kramer et~al\mbox{.}(2026)]%
        {kramer2026causal}
\bibfield{author}{\bibinfo{person}{Patrick Kramer}, \bibinfo{person}{Edward~H Kennedy}, {and} \bibinfo{person}{Isaac~M Opper}.} \bibinfo{year}{2026}\natexlab{}.
\newblock \showarticletitle{Causal inference with high-dimensional treatments}.
\newblock \bibinfo{journal}{\emph{arXiv:2602.21423}} (\bibinfo{year}{2026}).
\newblock


\bibitem[Lee et~al\mbox{.}(2022)]%
        {Lee.2022}
\bibfield{author}{\bibinfo{person}{Katherine Lee}, \bibinfo{person}{Daphne Ippolito}, \bibinfo{person}{Andrew Nystrom}, \bibinfo{person}{Chiyuan Zhang}, \bibinfo{person}{Douglas Eck}, \bibinfo{person}{Chris Callison-Burch}, {and} \bibinfo{person}{Nicholas Carlini}.} \bibinfo{year}{2022}\natexlab{}.
\newblock \showarticletitle{Deduplicating training data makes language models better}. In \bibinfo{booktitle}{\emph{ACL}}.
\newblock


\bibitem[Leung et~al\mbox{.}(2026)]%
        {Leung.2026}
\bibfield{author}{\bibinfo{person}{Kin~Kwan Leung}, \bibinfo{person}{Mouloud Belbahri}, \bibinfo{person}{Yi Sui}, \bibinfo{person}{Alex Labach}, \bibinfo{person}{Xueying Zhang}, \bibinfo{person}{Stephen~Anthony Rose}, {and} \bibinfo{person}{Jesse~C. Cresswell}.} \bibinfo{year}{2026}\natexlab{}.
\newblock \showarticletitle{Classifying and addressing the diversity of errors in retrieval-augmented generation systems}. In \bibinfo{booktitle}{\emph{EACL}}.
\newblock


\bibitem[Lewis et~al\mbox{.}(2020)]%
        {Lewis.2020}
\bibfield{author}{\bibinfo{person}{Patrick Lewis}, \bibinfo{person}{Ethan Perez}, \bibinfo{person}{Aleksandra Piktus}, \bibinfo{person}{Fabio Petroni}, \bibinfo{person}{Vladimir Karpukhin}, \bibinfo{person}{Naman Goyal}, \bibinfo{person}{Heinrich K\"{u}ttler}, \bibinfo{person}{Mike Lewis}, \bibinfo{person}{Wen-tau Yih}, \bibinfo{person}{Tim Rockt\"{a}schel}, \bibinfo{person}{Sebastian Riedel}, {and} \bibinfo{person}{Douwe Kiela}.} \bibinfo{year}{2020}\natexlab{}.
\newblock \showarticletitle{Retrieval-augmented generation for knowledge-intensive NLP tasks}. In \bibinfo{booktitle}{\emph{NeurIPS}}.
\newblock


\bibitem[Li and Li(2025)]%
        {Li.2025}
\bibfield{author}{\bibinfo{person}{Xiudi Li} {and} \bibinfo{person}{Sijia Li}.} \bibinfo{year}{2025}\natexlab{}.
\newblock \showarticletitle{Efficient inference for covariate-adjusted Bradley-Terry model with covariate shift}.
\newblock \bibinfo{journal}{\emph{arXiv:2503.18256}} (\bibinfo{year}{2025}).
\newblock


\bibitem[Mire et~al\mbox{.}(2025)]%
        {Mire.2025}
\bibfield{author}{\bibinfo{person}{Joel Mire}, \bibinfo{person}{Zubin~Trivadi Aysola}, \bibinfo{person}{Daniel Chechelnitsky}, \bibinfo{person}{Nicholas Deas}, \bibinfo{person}{Chrysoula Zerva}, {and} \bibinfo{person}{Maarten Sap}.} \bibinfo{year}{2025}\natexlab{}.
\newblock \showarticletitle{Rejected dialects: Biases against {A}frican {A}merican Language in reward models}. In \bibinfo{booktitle}{\emph{Findings of the ACL: NAACL 2025}}.
\newblock


\bibitem[Miroyan et~al\mbox{.}(2025)]%
        {Miroyan.2025}
\bibfield{author}{\bibinfo{person}{Mihran Miroyan}, \bibinfo{person}{Tsung-Han Wu}, \bibinfo{person}{Logan King}, \bibinfo{person}{Tianle Li}, \bibinfo{person}{Jiayi Pan}, \bibinfo{person}{Xinyan Hu}, \bibinfo{person}{Wei-Lin Chiang}, \bibinfo{person}{Anastasios~N. Angelopoulos}, \bibinfo{person}{Trevor Darrell}, \bibinfo{person}{Narges Norouzi}, {and} \bibinfo{person}{Joseph~E. Gonzalez}.} \bibinfo{year}{2025}\natexlab{}.
\newblock \showarticletitle{Search {{Arena}}: analyzing search-augmented {{LLMs}}}.
\newblock \bibinfo{journal}{\emph{arXiv:2506.05334}} (\bibinfo{year}{2025}).
\newblock


\bibitem[Oprescu et~al\mbox{.}(2023)]%
        {Oprescu.2023}
\bibfield{author}{\bibinfo{person}{Miruna Oprescu}, \bibinfo{person}{Jacob Dorn}, \bibinfo{person}{Marah Ghoummaid}, \bibinfo{person}{Andrew Jesson}, \bibinfo{person}{Nathan Kallus}, {and} \bibinfo{person}{Uri Shalit}.} \bibinfo{year}{2023}\natexlab{}.
\newblock \showarticletitle{B-learner: Quasi-oracle bounds on heterogeneous causal effects under hidden confounding}. In \bibinfo{booktitle}{\emph{ICML}}.
\newblock


\bibitem[Ouyang et~al\mbox{.}(2022)]%
        {Ouyang.2022}
\bibfield{author}{\bibinfo{person}{Long Ouyang}, \bibinfo{person}{Jeff Wu}, \bibinfo{person}{Xu Jiang}, \bibinfo{person}{Diogo Almeida}, \bibinfo{person}{Carroll~L Wainwright}, \bibinfo{person}{Pamela Mishkin}, \bibinfo{person}{Chong Zhang}, \bibinfo{person}{Sandhini Agarwal}, \bibinfo{person}{Katarina Slama}, \bibinfo{person}{Alex Ray}, \bibinfo{person}{John Schulman}, \bibinfo{person}{Jacob Hilton}, \bibinfo{person}{Fraser Kelton}, \bibinfo{person}{Luke Miller}, \bibinfo{person}{Maddie Simens}, \bibinfo{person}{Amanda Askell}, \bibinfo{person}{Peter Welinder}, \bibinfo{person}{Paul Christiano}, \bibinfo{person}{Jan Leike}, {and} \bibinfo{person}{Ryan Lowe}.} \bibinfo{year}{2022}\natexlab{}.
\newblock \showarticletitle{Training language models to follow instructions with human feedback}. In \bibinfo{booktitle}{\emph{{{NeurIPS}}}}.
\newblock


\bibitem[Packer et~al\mbox{.}(2023)]%
        {packer2023memgpt}
\bibfield{author}{\bibinfo{person}{Charles Packer}, \bibinfo{person}{Vivian Fang}, \bibinfo{person}{Shishir\_G Patil}, \bibinfo{person}{Kevin Lin}, \bibinfo{person}{Sarah Wooders}, {and} \bibinfo{person}{Joseph\_E Gonzalez}.} \bibinfo{year}{2023}\natexlab{}.
\newblock \showarticletitle{{MemGPT}: towards LLMs as operating systems.}
\newblock  (\bibinfo{year}{2023}).
\newblock


\bibitem[Park et~al\mbox{.}(2024)]%
        {park2024offsetbias}
\bibfield{author}{\bibinfo{person}{Junsoo Park}, \bibinfo{person}{Seungyeon Jwa}, \bibinfo{person}{Ren Meiying}, \bibinfo{person}{Daeyoung Kim}, {and} \bibinfo{person}{Sanghyuk Choi}.} \bibinfo{year}{2024}\natexlab{}.
\newblock \showarticletitle{{O}ffset{B}ias: Leveraging debiased data for tuning evaluators}. In \bibinfo{booktitle}{\emph{Findings of the ACL: EMNLP 2024}}.
\newblock


\bibitem[Park et~al\mbox{.}(2023)]%
        {park2023generative}
\bibfield{author}{\bibinfo{person}{Joon~Sung Park}, \bibinfo{person}{Joseph O'Brien}, \bibinfo{person}{Carrie~Jun Cai}, \bibinfo{person}{Meredith~Ringel Morris}, \bibinfo{person}{Percy Liang}, {and} \bibinfo{person}{Michael~S Bernstein}.} \bibinfo{year}{2023}\natexlab{}.
\newblock \showarticletitle{Generative agents: Interactive simulacra of human behavior}. In \bibinfo{booktitle}{\emph{Annual ACM Symposium on User Interface Software and Technology}}.
\newblock


\bibitem[Pearl(2009)]%
        {Pearl.2009}
\bibfield{author}{\bibinfo{person}{Judea Pearl}.} \bibinfo{year}{2009}\natexlab{}.
\newblock \bibinfo{booktitle}{\emph{Causality}}.
\newblock \bibinfo{publisher}{{Cambridge University Press}}, \bibinfo{address}{New York City}.
\newblock
\showISBNx{9780521895606}


\bibitem[Rafailov et~al\mbox{.}(2023)]%
        {Rafailov.2023}
\bibfield{author}{\bibinfo{person}{Rafael Rafailov}, \bibinfo{person}{Archit Sharma}, \bibinfo{person}{Eric Mitchell}, \bibinfo{person}{Stefano Ermon}, \bibinfo{person}{Christopher~D Manning}, {and} \bibinfo{person}{Chelsea Finn}.} \bibinfo{year}{2023}\natexlab{}.
\newblock \showarticletitle{Direct preference optimization: {{Your}} language model is secretly a reward model}. In \bibinfo{booktitle}{\emph{{{NeurIPS}}}}.
\newblock


\bibitem[Robins et~al\mbox{.}(1994)]%
        {Robins.1994}
\bibfield{author}{\bibinfo{person}{James~M. Robins}, \bibinfo{person}{Andrea Rotnitzky}, {and} \bibinfo{person}{Lue~Ping Zhao}.} \bibinfo{year}{1994}\natexlab{}.
\newblock \showarticletitle{Estimation of regression coefficients when some regressors are not always observed}.
\newblock \bibinfo{journal}{\emph{Journal of the American Statistical Association}} \bibinfo{volume}{89}, \bibinfo{number}{427} (\bibinfo{year}{1994}), \bibinfo{pages}{846--866}.
\newblock


\bibitem[Rubin(1974)]%
        {Rubin.1974}
\bibfield{author}{\bibinfo{person}{Donald~B. Rubin}.} \bibinfo{year}{1974}\natexlab{}.
\newblock \showarticletitle{Estimating causal effects of treatments in randomized and nonrandomized studies}.
\newblock \bibinfo{journal}{\emph{Journal of Educational Psychology}} \bibinfo{volume}{66}, \bibinfo{number}{5} (\bibinfo{year}{1974}), \bibinfo{pages}{688--701}.
\newblock
\showISSN{0022-0663}


\bibitem[Saito and Joachims(2022)]%
        {saito2022off}
\bibfield{author}{\bibinfo{person}{Yuta Saito} {and} \bibinfo{person}{Thorsten Joachims}.} \bibinfo{year}{2022}\natexlab{}.
\newblock \showarticletitle{Off-policy evaluation for large action spaces via embeddings}. In \bibinfo{booktitle}{\emph{ICML}}.
\newblock


\bibitem[Saito et~al\mbox{.}(2023)]%
        {saito2023off}
\bibfield{author}{\bibinfo{person}{Yuta Saito}, \bibinfo{person}{Qingyang Ren}, {and} \bibinfo{person}{Thorsten Joachims}.} \bibinfo{year}{2023}\natexlab{}.
\newblock \showarticletitle{Off-policy evaluation for large action spaces via conjunct effect modeling}. In \bibinfo{booktitle}{\emph{ICML}}.
\newblock


\bibitem[Saito et~al\mbox{.}(2025)]%
        {saitopotec2025}
\bibfield{author}{\bibinfo{person}{Yuta Saito}, \bibinfo{person}{Jihan Yao}, {and} \bibinfo{person}{Thorsten Joachims}.} \bibinfo{year}{2025}\natexlab{}.
\newblock \showarticletitle{POTEC: Off-policy contextual bandits for large action spaces via policy decomposition}. In \bibinfo{booktitle}{\emph{ICLR}}.
\newblock


\bibitem[Sap et~al\mbox{.}(2019)]%
        {Sap.2019}
\bibfield{author}{\bibinfo{person}{Maarten Sap}, \bibinfo{person}{Dallas Card}, \bibinfo{person}{Saadia Gabriel}, \bibinfo{person}{Yejin Choi}, {and} \bibinfo{person}{Noah~A. Smith}.} \bibinfo{year}{2019}\natexlab{}.
\newblock \showarticletitle{The risk of racial bias in hate speech detection}. In \bibinfo{booktitle}{\emph{ACL}}.
\newblock


\bibitem[Sawarni et~al\mbox{.}(2025)]%
        {Sawarni.2025}
\bibfield{author}{\bibinfo{person}{Ayush Sawarni}, \bibinfo{person}{Sahasrajit Sarmasarkar}, {and} \bibinfo{person}{Vasilis Syrgkanis}.} \bibinfo{year}{2025}\natexlab{}.
\newblock \showarticletitle{Preference learning with response time: Robust losses and guarantees}. In \bibinfo{booktitle}{\emph{NeurIPS}}.
\newblock


\bibitem[Schick et~al\mbox{.}(2023)]%
        {schick2023toolformer}
\bibfield{author}{\bibinfo{person}{Timo Schick}, \bibinfo{person}{Jane Dwivedi-Yu}, \bibinfo{person}{Roberto Dessi}, \bibinfo{person}{Roberta Raileanu}, \bibinfo{person}{Maria Lomeli}, \bibinfo{person}{Eric Hambro}, \bibinfo{person}{Luke Zettlemoyer}, \bibinfo{person}{Nicola Cancedda}, {and} \bibinfo{person}{Thomas Scialom}.} \bibinfo{year}{2023}\natexlab{}.
\newblock \showarticletitle{Toolformer: Language Models can teach themselves to use tools}. In \bibinfo{booktitle}{\emph{NeurIPS}}.
\newblock


\bibitem[Schr{\"o}der et~al\mbox{.}(2026)]%
        {Schroeder.2026}
\bibfield{author}{\bibinfo{person}{Maresa Schr{\"o}der}, \bibinfo{person}{Pascal Janetzky}, \bibinfo{person}{Michael Klar}, {and} \bibinfo{person}{Stefan Feuerriegel}.} \bibinfo{year}{2026}\natexlab{}.
\newblock \showarticletitle{{OrthoBO:} Orthogonal {B}ayesian hyperparameter optimization}.
\newblock \bibinfo{journal}{\emph{arXiv:2605.06454}} (\bibinfo{year}{2026}).
\newblock


\bibitem[Shalit et~al\mbox{.}(2017)]%
        {shalit.2017estimating}
\bibfield{author}{\bibinfo{person}{Uri Shalit}, \bibinfo{person}{Fredrik~D Johansson}, {and} \bibinfo{person}{David Sontag}.} \bibinfo{year}{2017}\natexlab{}.
\newblock \showarticletitle{Estimating individual treatment effect: Generalization bounds and algorithms}. In \bibinfo{booktitle}{\emph{ICML}}.
\newblock


\bibitem[Shinn et~al\mbox{.}(2023)]%
        {shinn2023reflexion}
\bibfield{author}{\bibinfo{person}{Noah Shinn}, \bibinfo{person}{Federico Cassano}, \bibinfo{person}{Ashwin Gopinath}, \bibinfo{person}{Karthik Narasimhan}, {and} \bibinfo{person}{Shunyu Yao}.} \bibinfo{year}{2023}\natexlab{}.
\newblock \showarticletitle{Reflexion: Language agents with verbal reinforcement learning}. In \bibinfo{booktitle}{\emph{NeurIPS}}.
\newblock


\bibitem[Spokoiny(2025)]%
        {Spokoiny.2025}
\bibfield{author}{\bibinfo{person}{Vladimir Spokoiny}.} \bibinfo{year}{2025}\natexlab{}.
\newblock \showarticletitle{Semiparametric plug-in estimation, sup-norm risk bounds, marginal optimization, and inference in {{BTL}} model}.
\newblock \bibinfo{journal}{\emph{arXiv:2503.15045}} (\bibinfo{year}{2025}).
\newblock


\bibitem[Touvron et~al\mbox{.}(2023)]%
        {touvron2023llama}
\bibfield{author}{\bibinfo{person}{Hugo Touvron}, \bibinfo{person}{Thibaut Lavril}, \bibinfo{person}{Gautier Izacard}, \bibinfo{person}{Xavier Martinet}, \bibinfo{person}{Marie-Anne Lachaux}, \bibinfo{person}{Timoth{\'e}e Lacroix}, \bibinfo{person}{Baptiste Rozi{\`e}re}, \bibinfo{person}{Naman Goyal}, \bibinfo{person}{Eric Hambro}, \bibinfo{person}{Faisal Azhar}, {et~al\mbox{.}}} \bibinfo{year}{2023}\natexlab{}.
\newblock \showarticletitle{Llama: Open and efficient foundation language models}.
\newblock \bibinfo{journal}{\emph{arXiv preprint arXiv:2302.13971}} (\bibinfo{year}{2023}).
\newblock


\bibitem[Tsiourvas et~al\mbox{.}(2025)]%
        {Tsiourvas.2025}
\bibfield{author}{\bibinfo{person}{Asterios Tsiourvas}, \bibinfo{person}{Wei Sun}, {and} \bibinfo{person}{Georgia Perakis}.} \bibinfo{year}{2025}\natexlab{}.
\newblock \showarticletitle{Causal LLM routing: End-to-end regret minimization from observational data}. In \bibinfo{booktitle}{\emph{NeurIPS}}.
\newblock


\bibitem[Ulichney and Coston(2025)]%
        {Ulichney.2025}
\bibfield{author}{\bibinfo{person}{Annie~S. Ulichney} {and} \bibinfo{person}{Amanda~Lee Coston}.} \bibinfo{year}{2025}\natexlab{}.
\newblock \showarticletitle{Double machine learning evaluation under distribution shift and selection bias}. In \bibinfo{booktitle}{\emph{{{NeurIPS}} 2025 {{Workshop}}: {{Reliable ML}} from {{Unreliable Data}}}}.
\newblock


\bibitem[van~der Laan and Van Der~Laan(2026)]%
        {van2026calibeating}
\bibfield{author}{\bibinfo{person}{Lars van~der Laan} {and} \bibinfo{person}{Mark Van Der~Laan}.} \bibinfo{year}{2026}\natexlab{}.
\newblock \showarticletitle{Calibeating prediction-powered inference}.
\newblock \bibinfo{journal}{\emph{arXiv:2604.21260}} (\bibinfo{year}{2026}).
\newblock


\bibitem[{van der Laan} and Rubin(2006)]%
        {vanderLaan.2006}
\bibfield{author}{\bibinfo{person}{Mark~J. {van der Laan}} {and} \bibinfo{person}{Donald~B. Rubin}.} \bibinfo{year}{2006}\natexlab{}.
\newblock \showarticletitle{Targeted maximum likelihood learning}.
\newblock \bibinfo{journal}{\emph{The International Journal of Biostatistics}} \bibinfo{volume}{2}, \bibinfo{number}{1} (\bibinfo{year}{2006}).
\newblock


\bibitem[{van der Vaart}(1998)]%
        {vanderVaart.1998}
\bibfield{author}{\bibinfo{person}{Aart {van der Vaart}}.} \bibinfo{year}{1998}\natexlab{}.
\newblock \bibinfo{booktitle}{\emph{Asymptotic statistics}}.
\newblock \bibinfo{publisher}{Cambridge University Press}, \bibinfo{address}{Cambridge}.
\newblock
\showISBNx{0-521-49603-9}


\bibitem[Veitch et~al\mbox{.}(2020)]%
        {Veitch.2020}
\bibfield{author}{\bibinfo{person}{Victor Veitch}, \bibinfo{person}{Dhanya Sridhar}, {and} \bibinfo{person}{David Blei}.} \bibinfo{year}{2020}\natexlab{}.
\newblock \showarticletitle{Adapting text embeddings for causal inference}. In \bibinfo{booktitle}{\emph{UAI}}.
\newblock


\bibitem[Wang et~al\mbox{.}(2024b)]%
        {wang2024voyager}
\bibfield{author}{\bibinfo{person}{Guanzhi Wang}, \bibinfo{person}{Yuqi Xie}, \bibinfo{person}{Yunfan Jiang}, \bibinfo{person}{Ajay Mandlekar}, \bibinfo{person}{Chaowei Xiao}, \bibinfo{person}{Yuke Zhu}, \bibinfo{person}{Linxi Fan}, {and} \bibinfo{person}{Anima Anandkumar}.} \bibinfo{year}{2024}\natexlab{b}.
\newblock \showarticletitle{Voyager: An open-ended embodied agent with large language models}.
\newblock \bibinfo{journal}{\emph{TMLR}} (\bibinfo{year}{2024}).
\newblock


\bibitem[Wang et~al\mbox{.}(2024a)]%
        {wang-etal-2024-large-language-models-fair}
\bibfield{author}{\bibinfo{person}{Peiyi Wang}, \bibinfo{person}{Lei Li}, \bibinfo{person}{Liang Chen}, \bibinfo{person}{Zefan Cai}, \bibinfo{person}{Dawei Zhu}, \bibinfo{person}{Binghuai Lin}, \bibinfo{person}{Yunbo Cao}, \bibinfo{person}{Lingpeng Kong}, \bibinfo{person}{Qi Liu}, \bibinfo{person}{Tianyu Liu}, {and} \bibinfo{person}{Zhifang Sui}.} \bibinfo{year}{2024}\natexlab{a}.
\newblock \showarticletitle{Large language models are not fair evaluators}. In \bibinfo{booktitle}{\emph{ACL}}.
\newblock


\bibitem[Wu et~al\mbox{.}(2026)]%
        {Wu.2026}
\bibfield{author}{\bibinfo{person}{Shirley Wu}, \bibinfo{person}{Parth Sarthi}, \bibinfo{person}{Shiyu Zhao}, \bibinfo{person}{Aaron Lee}, \bibinfo{person}{Herumb Shandilya}, \bibinfo{person}{Adrian~Mladenic Grobelnik}, \bibinfo{person}{Nurendra Choudhary}, \bibinfo{person}{Edward~W Huang}, \bibinfo{person}{Karthik Subbian}, \bibinfo{person}{Linjun Zhang}, \bibinfo{person}{Diyi Yang}, \bibinfo{person}{James Zou}, {and} \bibinfo{person}{Jure Leskovec}.} \bibinfo{year}{2026}\natexlab{}.
\newblock \showarticletitle{Optimas: Optimizing compound {AI} systems with globally aligned local rewards}. In \bibinfo{booktitle}{\emph{ICLR}}.
\newblock


\bibitem[Xia et~al\mbox{.}(2024)]%
        {Xia.2024}
\bibfield{author}{\bibinfo{person}{Yu Xia}, \bibinfo{person}{Tong Yu}, \bibinfo{person}{Zhankui He}, \bibinfo{person}{Handong Zhao}, \bibinfo{person}{Julian McAuley}, {and} \bibinfo{person}{Shuai Li}.} \bibinfo{year}{2024}\natexlab{}.
\newblock \showarticletitle{Aligning as debiasing: {{Causality-aware}} Alignment via reinforcement learning with interventional feedback}. In \bibinfo{booktitle}{\emph{{{NAACL}}}}.
\newblock


\bibitem[Xu et~al\mbox{.}(2025b)]%
        {Xu.2025b}
\bibfield{author}{\bibinfo{person}{Erhan Xu}, \bibinfo{person}{Kai Ye}, \bibinfo{person}{Hongyi Zhou}, \bibinfo{person}{Luhan Zhu}, \bibinfo{person}{Francesco Quinzan}, {and} \bibinfo{person}{Chengchun Shi}.} \bibinfo{year}{2025}\natexlab{b}.
\newblock \showarticletitle{Doubly robust alignment for large language models}. In \bibinfo{booktitle}{\emph{{{NeurIPS}}}}.
\newblock


\bibitem[Xu et~al\mbox{.}(2025a)]%
        {Xu.2025}
\bibfield{author}{\bibinfo{person}{Yi Xu}, \bibinfo{person}{Laura Ruis}, \bibinfo{person}{Tim Rockt{\"a}schel}, {and} \bibinfo{person}{Robert Kirk}.} \bibinfo{year}{2025}\natexlab{a}.
\newblock \showarticletitle{Investigating non-transitivity in {LLM-as-a-judge}}. In \bibinfo{booktitle}{\emph{ICML}}.
\newblock


\bibitem[Yao et~al\mbox{.}(2023)]%
        {Yao.2023}
\bibfield{author}{\bibinfo{person}{Shunyu Yao}, \bibinfo{person}{Jeffrey Zhao}, \bibinfo{person}{Dian Yu}, \bibinfo{person}{Nan Du}, \bibinfo{person}{Izhak Shafran}, \bibinfo{person}{Karthik~R Narasimhan}, {and} \bibinfo{person}{Yuan Cao}.} \bibinfo{year}{2023}\natexlab{}.
\newblock \showarticletitle{{ReAct:} Synergizing reasoning and acting in language models}. In \bibinfo{booktitle}{\emph{ICLR}}.
\newblock


\bibitem[Ze{\v{c}}evi{\'c} et~al\mbox{.}(2023)]%
        {zevcevic2023causal}
\bibfield{author}{\bibinfo{person}{Matej Ze{\v{c}}evi{\'c}}, \bibinfo{person}{Moritz Willig}, \bibinfo{person}{Devendra~Singh Dhami}, {and} \bibinfo{person}{Kristian Kersting}.} \bibinfo{year}{2023}\natexlab{}.
\newblock \showarticletitle{Causal parrots: Large language models may talk causality but are not causal}.
\newblock \bibinfo{journal}{\emph{TMLR}} (\bibinfo{year}{2023}).
\newblock


\bibitem[Zhang and Hardt(2024)]%
        {Zhang.2024}
\bibfield{author}{\bibinfo{person}{Guanhua Zhang} {and} \bibinfo{person}{Moritz Hardt}.} \bibinfo{year}{2024}\natexlab{}.
\newblock \showarticletitle{Inherent trade-offs between diversity and stability in multi-task benchmarks}. In \bibinfo{booktitle}{\emph{{{ICML}}}}.
\newblock


\bibitem[Zhang et~al\mbox{.}(2026)]%
        {Zhang.2025}
\bibfield{author}{\bibinfo{person}{Yichi Zhang}, \bibinfo{person}{Fangzheng Xie}, \bibinfo{person}{Shu Yang}, {and} \bibinfo{person}{Chong Wu}.} \bibinfo{year}{2026}\natexlab{}.
\newblock \showarticletitle{{Meta-Router}: Bridging gold-standard and preference-based evaluations in {LLM} routing}. In \bibinfo{booktitle}{\emph{ICLR}}.
\newblock


\bibitem[Zhao et~al\mbox{.}(2023)]%
        {zhao2023survey}
\bibfield{author}{\bibinfo{person}{Wayne~Xin Zhao}, \bibinfo{person}{Kun Zhou}, \bibinfo{person}{Junyi Li}, \bibinfo{person}{Tianyi Tang}, \bibinfo{person}{Xiaolei Wang}, \bibinfo{person}{Yupeng Hou}, \bibinfo{person}{Yingqian Min}, \bibinfo{person}{Beichen Zhang}, \bibinfo{person}{Junjie Zhang}, \bibinfo{person}{Zican Dong}, {et~al\mbox{.}}} \bibinfo{year}{2023}\natexlab{}.
\newblock \showarticletitle{A survey of large language models}.
\newblock \bibinfo{journal}{\emph{arXiv:2303.18223}} (\bibinfo{year}{2023}).
\newblock


\bibitem[Zhou et~al\mbox{.}(2024)]%
        {zhou2024webarena}
\bibfield{author}{\bibinfo{person}{Shuyan Zhou}, \bibinfo{person}{Frank~F. Xu}, \bibinfo{person}{Hao Zhu}, \bibinfo{person}{Xuhui Zhou}, \bibinfo{person}{Robert Lo}, \bibinfo{person}{Abishek Sridhar}, \bibinfo{person}{Xianyi Cheng}, \bibinfo{person}{Tianyue Ou}, \bibinfo{person}{Yonatan Bisk}, \bibinfo{person}{Daniel Fried}, \bibinfo{person}{Uri Alon}, {and} \bibinfo{person}{Graham Neubig}.} \bibinfo{year}{2024}\natexlab{}.
\newblock \showarticletitle{{WebArena}: A realistic web environment for building autonomous agents}. In \bibinfo{booktitle}{\emph{ICLR}}.
\newblock


\bibitem[Zrnic and Cand{\`e}s(2024)]%
        {Zrnic.2024}
\bibfield{author}{\bibinfo{person}{Tijana Zrnic} {and} \bibinfo{person}{Emmanuel~J. Cand{\`e}s}.} \bibinfo{year}{2024}\natexlab{}.
\newblock \showarticletitle{Cross-prediction-powered inference}.
\newblock \bibinfo{journal}{\emph{Proceedings of the National Academy of Sciences}} \bibinfo{volume}{121}, \bibinfo{number}{15} (\bibinfo{year}{2024}), \bibinfo{pages}{e2322083121}.
\newblock


\end{thebibliography}

\appendix



\end{document}